\documentclass[10pt,twocolumn,letterpaper]{article}

\usepackage{iccv}
\usepackage{times}
\usepackage{epsfig}
\usepackage{graphicx}
\usepackage{amsmath}
\usepackage{amssymb}
\usepackage{booktabs}
\usepackage{multirow}
\usepackage{enumitem}
\usepackage[table]{xcolor}
\usepackage{color}
\usepackage{xcolor}
\usepackage{ctable}

\usepackage[pagebackref=true,breaklinks=true,letterpaper=true,colorlinks,bookmarks=false]{hyperref}

\usepackage{colortbl}
\def\mycolor{\cellcolor[rgb]{0.90,0.90,0.90}}
\def \robcolor{\textcolor[rgb]{0.1,0.1,0.8}} 
\def \redcolor{\textcolor[rgb]{0.8,0.1,0.1}} 
\usepackage[capitalize]{cleveref}
\definecolor{Gray}{gray}{0.92}
\crefname{section}{Sec.}{Secs.}
\Crefname{section}{Section}{Sections}
\Crefname{table}{Table}{Tables}
\crefname{table}{Tab.}{Tabs.}

\usepackage{amssymb}
\usepackage{pifont}
\newcommand{\cmark}{\ding{51}}%
\newcommand{\xmark}{\ding{55}}%

\iccvfinalcopy 

\def\iccvPaperID{****} 

\ificcvfinal\fi

\newcommand{\ourMethod}{FBMNet}

\begin{document}

\title{Multi-Modal 3D Object Detection by Box Matching}


\author{
Zhe~Liu\textsuperscript{\rm 1}
\quad Xiaoqing~Ye\textsuperscript{\rm 2} 
\quad Zhikang~Zou\textsuperscript{\rm 2}
\quad Xinwei~He\textsuperscript{\rm 3} \\
\quad Xiao~Tan\textsuperscript{\rm 2} 
\quad Errui~Ding\textsuperscript{\rm 2} 
\quad Jingdong~Wang\textsuperscript{\rm 2} 
\quad Xiang Bai\textsuperscript{\rm 1}\thanks{Corresponding Author.} \\
{\textsuperscript{\rm 1}Huazhong University of Science and Technology, \textsuperscript{\rm 2}Baidu Inc., China}\\
{\textsuperscript{\rm 3}Huazhong Agricultural University} \\
\small{\texttt{\{zheliu1994,xbai\}@hust.edu.cn, eriche.hust@gmail.com
}} \\
\small{\texttt{\{yexiaoqing,zouzhikang,tanxiao01,dingerrui,wangjingdong\}@baidu.com}
}
}

\maketitle
\ificcvfinal\thispagestyle{empty}\fi
\begin{abstract}
Multi-modal 3D object detection has received growing attention as the information from different sensors like LiDAR and cameras are complementary. 
Most fusion methods for 3D detection rely on an accurate alignment and calibration between 3D point clouds and RGB images. 
However, such an assumption is not reliable in a real-world self-driving system, as the alignment between different modalities is easily affected by asynchronous sensors and disturbed sensor placement. We propose a novel \textbf{F}usion network by \textbf{B}ox \textbf{M}atching (\textbf{FBMNet}) for multi-modal 3D detection, which provides an alternative way 
for cross-modal feature alignment by learning the correspondence at the bounding box level to free up the dependency of calibration during inference. With the learned assignments between 3D and 2D object proposals, 
the fusion for detection can be effectively performed by combing their ROI features. 
Extensive experiments on the nuScenes dataset demonstrate that our method is much more 
stable in dealing with challenging cases such as asynchronous sensors,  misaligned sensor placement, and degenerated camera images than existing fusion methods. We hope that our {\ourMethod} could provide an available solution to dealing with these challenging cases for safety in real autonomous driving scenarios. Codes will be publicly available at \url{https://github.com/happinesslz/FBMNet}.
\end{abstract}

\section{Introduction}
\label{sec:intro}

\begin{figure}[t]
	\centering
	\includegraphics[width=1.0\linewidth]{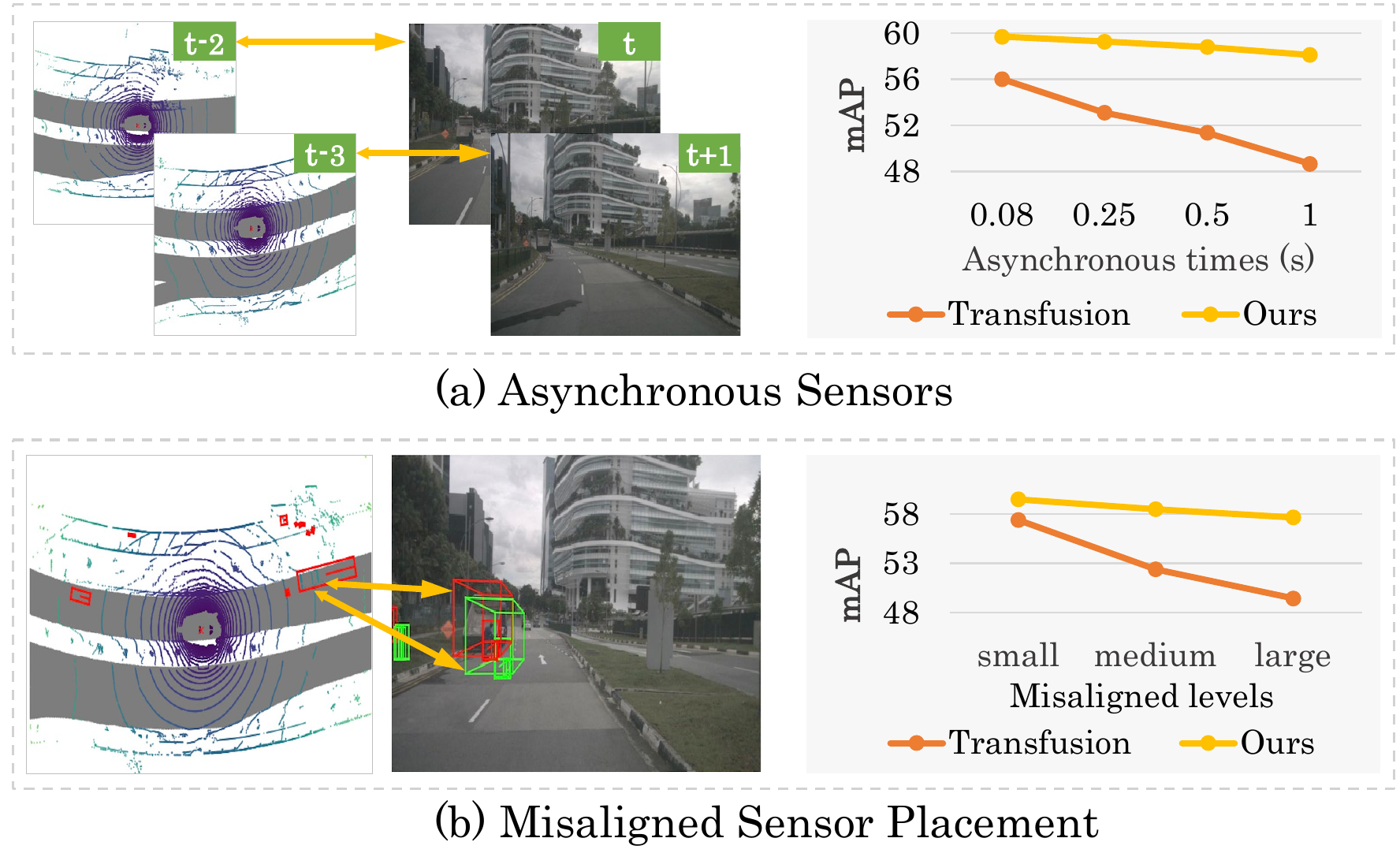}
	\caption{
(a) and (b) indicate the challenging cases caused by asynchronous sensors and misaligned sensor placement, respectively. The right graphs depict the corresponding results with the evaluation metric of mAP on nuScenes~\cite{caesar2020nuscenes} under these cases for both Transfusion~\cite{bai2022transfusion} and our box-matching framework. For a fair comparison, except the projection-based fusion and our box-matching fusion, we adopt the same network structure, including a pillar-based Transfusion-L~\cite{bai2022transfusion} as the LiDAR branch and a 2D detector Mask R-CNN~\cite{he2017mask} as the image branch.
		}\label{fig:intro}
  \vspace{-8pt}
\end{figure}

LiDARs and cameras are the popular sensors for 3D perception as the components of an auto-driving system~\cite{dosovitskiy2017carla,gaidon2016virtual,sun2020scalability,prakash2021multi}. Recently, there has been a trend to combine the information collected by LiDARs and cameras for 3D object detection, as camera images and LiDAR points often provide complementary information for identifying objects. Most approaches~\cite{sindagi2019mvx,vora2020pointpainting,chen2022focal,wang2021pointaugmenting} for multi-modal 3D detection focus on fusing the representation of camera images and LiDAR points at a network’s different stages, achieving superior performance over 3D detectors based on a single modality. Such methods benefit from an ideal assumption that the accurate projections between two modalities are given before the fusion process, while it is extremely difficult to obtain accurate calibration matrices from a real-world system due to the common cases such as sensor jitter, different frame rates of the sensors, etc. We argue that the calibration issue of associating the correspondence between LiDAR points and camera images has been more or less neglected by previous methods for multi-sensor fusion.

In general, most mainstream works for multi-sensor fusion heavily rely on accurate calibrations, which can be coarsely divided into three categories: input level, decision level, and feature level. Input-level fusion methods~\cite{vora2020pointpainting,wang2021pointaugmenting} combine LiDAR points with the semantic scores or features produced by an off-the-shelf image segmentation network in a point-wise manner, which require a point-to-image projection before the fusion operations. The decision-level method CLOCs~\cite{pang2020clocs} first performs 2D and 3D object detection based on camera images and lidar points respectively then associates the detected 2D and 3D bounding boxes according to the projection between two modalities in order to obtain the final results. Feature-level fusion methods~\cite{chen2022autoalignv2,li2022deepfusion,huang2020epnet} combine the features extracted from LiDAR points and camera images according to the point-wise or voxel-wise correspondence that is also established by a projection matrix. 

Although the above fusion methods have achieved promising progress in most cases, they may suffer from performance degradation due to two major challenges: 1) The temporal alignment between two modalities can not often be guaranteed due to the asynchronous sensors; 2) In an open environment, there often exists calibration errors caused by sensor jitter or fast motion.  As illustrated in Figure~\ref{fig:intro}, the detection performance of a recent robust fusion method TransFusion~\cite{bai2022transfusion} drops significantly when dealing with cases of asynchronous sensors and misaligned sensor placement. This phenomenon shows that the spatial and temporal alignments between different modalities are crucial for multi-modal object detection.

In this paper, we propose an unconventional multi-modal \textbf{F}usion network by \textbf{B}ox \textbf{M}atching~(\textbf{FBMNet}) for 3D detection without the prior alignments between LiDAR and camera sensors in the inference stage. Our motivation is to directly learn the correspondence between 2D and 3D object proposals, as the proposal alignments are much more robust than the point correspondences provided by projection matrices. Specifically, the proposed box matching is composed of two steps: 1) Given each view image, we first learn to select a set of 3D proposals that are relevant to it; 2) A learning-based matching algorithm is proposed to establish the correspondence between the selected 3D proposals and 2D proposals from each view image.
Dividing the associating process into two steps has the merits of both efficiency and efficacy. The first step aims at narrowing down the 2D proposal search space of specific cameras for each 3D proposal, based on which the second matching identifies the 2D-3D proposal-level association more simply. 
With the two-level matching, we easily obtain the associated proposal pairs and further fuse their ROI features without the calibration.
To the best of our knowledge, our work is the first to study the calibration-free multi-modal fusion for 3D detection. The proposed box-matching scheme has several desirable properties. It eliminates the need for accurate calibrations between the camera and LiDAR sensors in the inference stage, which is costly to obtain in practice. More importantly, it gives our method a high level of robustness to asynchronous and calibration issues  (shown in Figure~\ref{fig:intro}), which is extremely important in open-world driving scenarios. 
In summary, our contributions are as follows:
\begin{itemize}
\item We introduce a novel robust multi-modal fusion  framework named FBMNet for 3D detection by the mechanism of box matching, which frees up the  heavy  dependency of a projection matrix adopted for most existing multi-modal 3D detection methods. 
\vspace*{-2mm}
\item We propose a two-level matching process (\ie, view-level matching and proposal-level matching) to efficiently learn to associate 3D and 2D proposals in a \textit{coarse-to-fine} manner, requiring no calibrations in the inference stage and demonstrating robustness to asynchronous sensors, misaligned sensor placement, and even the degenerated camera images when compared with existing methods. 
\end{itemize}

\begin{figure*}[t]
	\centering
	\includegraphics[width=0.95\linewidth]{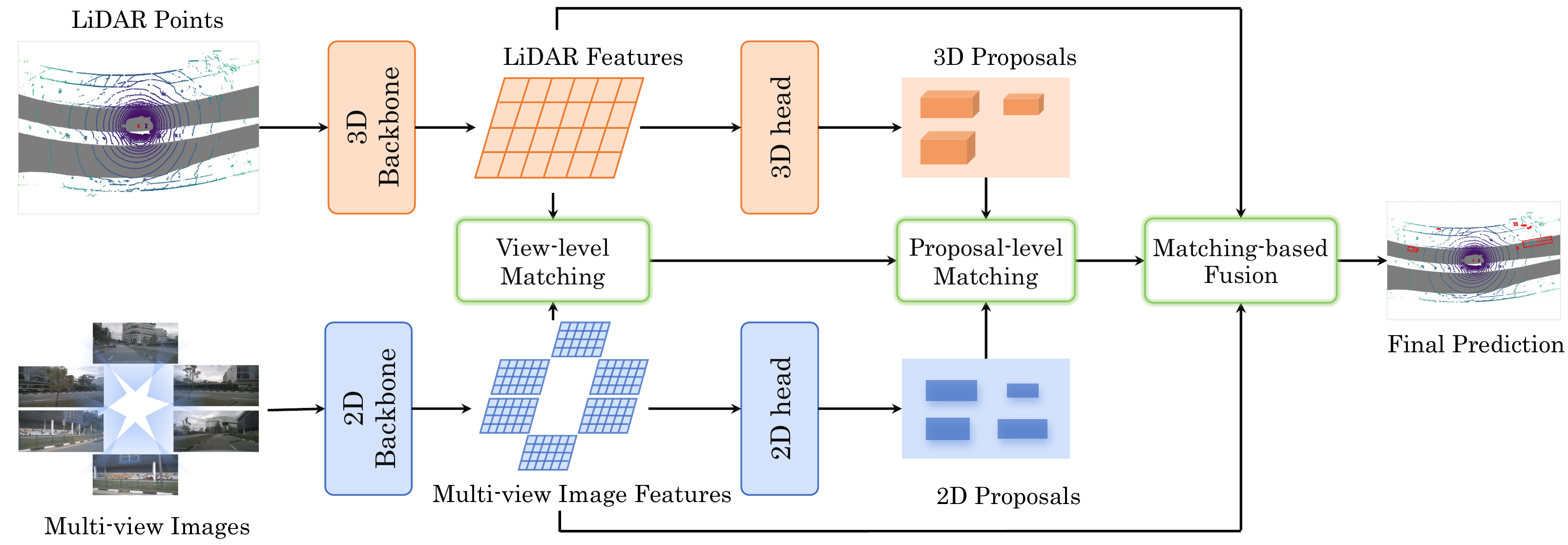}
	\caption{The framework of our \ourMethod{}. The image and LiDAR branches individually feed the image and LiDAR points into the 2D and 3D networks, producing 2D/3D modality-specific features and proposals. Note that the following mentioned 3D object proposal is initially generated by the LiDAR branch~(TransFusion-L~\cite{bai2022transfusion}). To free up the projection matrix, we first perform view-level matching by assigning 3D object proposals to the specified view image. Then, the proposal-level matching further establishes the proposal-wise correspondence between 3D and 2D proposals in the matched image view. After obtaining the proposal-level correspondences, we fuse the features with a matching-based fusion module for final 3D detection.}\label{fig:framework}
  \vspace{-8pt}
\end{figure*}

\section{Related Works}
\label{sec:related}


\noindent \textbf{LiDAR-only 3D Detection.} Existing LiDAR-only methods can be broadly grouped into three types: \textit{i.e.}, point-based, voxel-based and hybrid point-voxel based. Point-based detectors~\cite{yang2019std, shi2019pointrcnn, shi2020pointgnn, yang20203dssd, pan20213d, chen2022sasa,zhang2022not} directly extract sparse and irregular raw point cloud geometric and structure features by a PointNet-like point cloud network~\cite{qi2017pointnet,qi2017pointnet++} and then achieve 3D object localization by a detection head. To effectively deal with the irregular point cloud, voxel-based detectors~\cite{yan2018second, zhou2018voxelnet, ye2020hvnet, deng2021voxel, zheng2020ciassd, mao2021voxel, yin2021center,du2021ago,bai2022transfusion,guan2022m3detr,liu2020tanet} further divide 3D points into regular 3D voxel grids and implement 3D convolutional operations to extract 3D voxel features, which can be organized to a 2D BEV space by compressing them along the height dimension. Most recently, the hybrid point-voxel methods~\cite{shi2020pv,FromVoxelToPoint,sheng2021improving,mao2021pyramid,noh2021hvpr,xu2022behind}  combine the merits of the voxel-based methods by a 3D sparse convolution to efficiently extract voxel-wise feature and the point-based approaches by adopting set abstraction~\cite{qi2017pointnet++} to aggregate more fine-grained geometric features. In this paper, we choose the advanced voxel-based detector TransFusion-L~\cite{bai2022transfusion} as our LiDAR branch.

\noindent\textbf{Multi-modal 3D Detection.} Benefiting from the complementary nature of point clouds and images, multi-modal fusion for 3D detection~\cite{chen2017multi, chen2022futr3d, sindagi2019mvx, huang2020epnet, xie2020pi,chen2022autoalignv2,li2022unifying,liang2018deep,xu2021fusionpainting,jiao2022msmdfusion,liu2022epnet++} has attracted much attention. Existing fusion-based methods can be roughly divided into three categories: \ie, input-level, decision-level and feature-level. The input-level approaches~\cite{vora2020pointpainting, wang2021pointaugmenting, yin2021multimodal} combine the input LiDAR points with image information in a point-wise manner by a projection matrix before feeding into a LiDAR-only 3D detector. 
The decision-level methods~\cite{pang2020clocs,pang2022fast} first achieve 2D and 3D detection in respective modalities and then fuse them together to obtain the final 3D results. This fusion paradigm benefits from exploiting the geometric and semantic consistency of outputs from two modalities while maintaining the independence of the two modal networks. The feature-level fusion~\cite{chen2022futr3d, li2022deepfusion,sindagi2019mvx,liang2022bevfusion, chen2022focal,yang2022deepinteraction,li2022voxel,wu2022sparse,chen2022autoalign} approaches first establish the point-wise or voxel-wise correspondence between two modalities by the projection matrix and then enhance the LiDAR features with rich semantic image features extracted by a deep neural network. 
Despite the promising results, existing fusion-based methods are highly dependent on the precise calibration of data from different modalities, which is difficult to sustain in practical scenarios. TransFusion~\cite{bai2022transfusion} proposes a soft association mechanism by a cross-attention operation to alleviate this problem, however, the process of building a soft association needs to search the local region of interest based on the reference points with the help of the projection matrix for effective fusion. Different from the above methods,
in this paper, we introduce a novel  fusion pipeline, which learns multi-modal box matching to free up the heavy dependency of the projection matrix.  

\noindent\textbf{Robustness in 3D Object Detection.} Recently, more and more attention has payed attention to the robustness on 3D object detection. \cite{yu2022benchmarking} proposes an efficient robust training strategy to boost the robustness of the current fusion approaches for 3D object detection tasks.
Robo3D~\cite{kong2023robo3d} provides a density-insensitive training the
framework through a simple flexible voxelization strategy to enhance the model resiliency for LiDAR-only detectors or segmentors~\cite{zhu2021cylindrical,choy20194d}. RoboBEV~\cite{xie2023robobev} provides a comprehensive benchmark for the robustness of BEV algorithms~\cite{li2022bevformer,huang2021bevdet,liu2022petr}, which  includes the corruptions of Bright, Dark, Fog, Snow, Motion Blur, Color Quant, Camera Crash, and Frame Lost. Besides, ~\cite{dong2023benchmarking}
simulates some corruptions in autonomous driving systems caused by adverse weathers, sensor noises, etc. Compared with the above research, we try to provide an available solution for multi-modal 3D object detection through box matching to deal with the challenging cases, which include the spatial and temporal misalignment between camera and LiDAR sensors, and camera sensor failure.

\section{Method}
\label{sec:method}

In this section, we introduce our multi-modal \textbf{F}usion network by \textbf{B}ox \textbf{M}atching~(\textbf{FBMNet}) for multi-modal 3D object detection. 
The architecture of \ourMethod{} is presented in Figure~\ref{fig:framework}, which consists of four parts: the LiDAR branch, the image branch, the two-level matching part, and the matching-based fusion module.

The LiDAR branch takes the LiDAR point clouds as input,
and predicts the class-specific heatmap and only preserves the top-$N_{3d}$ candidates through the circular NMS~\cite{yin2021center}. Then, we follow TransFusion-L~\cite{bai2022transfusion} to generate the initial 3D object queries $F_{3d} \in \mathbb{R}^{N_{3d} \times C}$, where $C$ is the feature dimension of object queries. 
Given $N_v$  camera view images, the image branch can generate multi-view image features with the shape of ${N_v \times H \times W \times C_{im}}$  and the corresponding 2D object proposals, where $H$, $W$ and $C_{im}$ represent the height, width, and the channel dimension of the multi-view image feature. To keep the same channel dimension $C$ as 3D object queries of the LiDAR branch, we feed them to a convolution with the kernel of $3\times 3$ and obtain the final image feature map $F_{im} \in \mathbb{R}^{N_v \times H \times W \times C}$.


Based on the above two-steam outputs, we further propose a calibration-free fusion method by utilizing a two-level matching strategy, namely, view-level matching and proposal-level matching. Thanks to these two-level matching, we can obtain the proposal-wise correspondence between LiDAR points and camera images, and then implement the proposal-level fusion by the matching-based fusion module instead of adopting a projection matrix for most existing multi-modal methods. 

\begin{figure}[t!]
	\centering
	\includegraphics[width=1\linewidth]{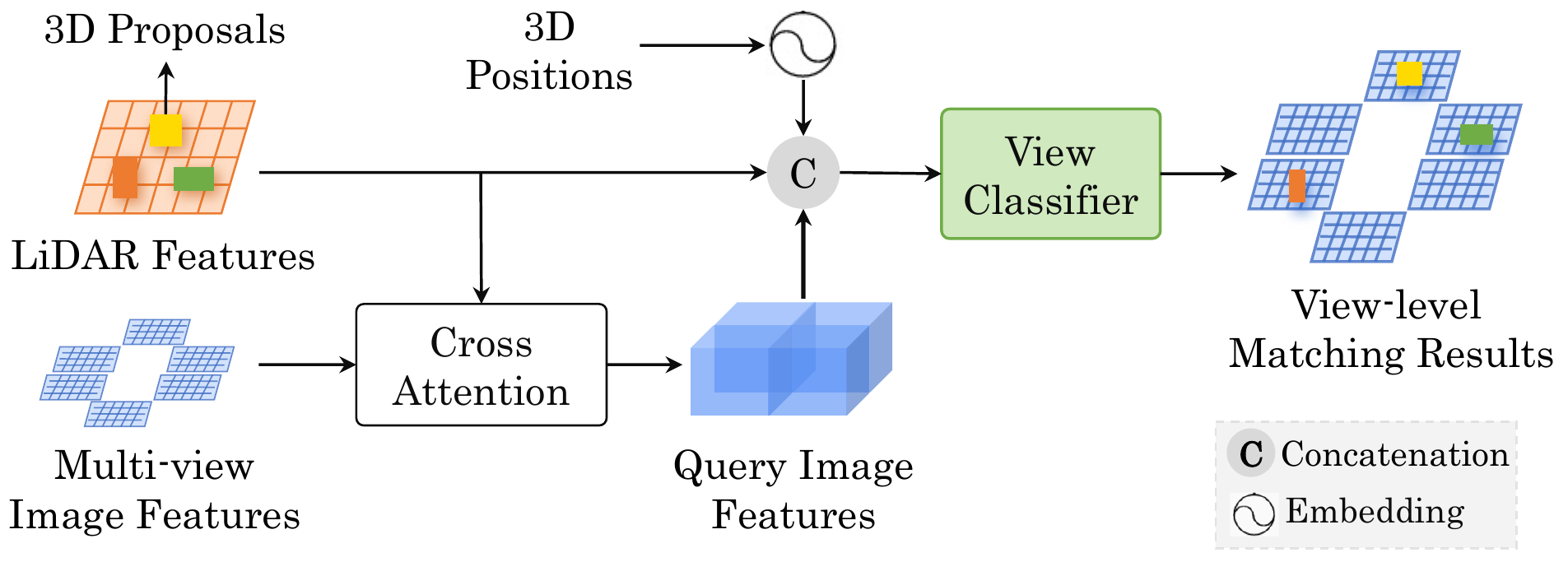}
	\caption{
The pipeline of View-level Matching. We adopt the view classifier to assign each 3D proposal to the corresponding image view without any calibration matrix.
		}\label{fig:view_matching}
  \vspace{-5pt}
\end{figure}

\subsection{View-level Matching}
\label{subsec:viewlevel}

Straightforwardly, an object can both be captured by LiDAR and one or more camera sensors at the same time when they have an overlapped field of view~(FOV). To fuse the complementary 2D and 3D features of the same object, we design a view-level matching strategy. In essence, view-level matching aims at associating each 3D proposal in LiDAR points with the corresponding image view. The pipeline of view-level matching is presented in Figure~\ref{fig:view_matching}.
Without the prior calibration matrix, we are unable to directly assign each 3D proposal to the target camera view.
To circumvent this issue, we formulate the process as a ($N_v+1$)-Category classification task considering the fact that the 3D proposal may fall outside of multiple views. 
In detail, to identify whether a 3D proposal is visible or not in a specified view image, we employ a ``view classifier'' to assign a 3D proposal to views. We collect three important cues: 1) The extracted 3D proposal features~(also called 3D object queries $F_{3d}$); 2) The interactive information of 3D proposal features with multi-view image features. 3) The 3D positional embedding features. 

To obtain the interactive cue of 3D proposal features with multi-view features, we first collapse the multi-view image feature $F_{im}$ along the height dimension inspired by \cite{bai2022transfusion,roddick2020predicting} to reduce the computation cost in the next process. The collapsing operation is based on the assumption that the vertical scan line of the image corresponds to a ray of the BEV map\cite{saha2022translating}. The collapsed image features can be denoted as $\mathit{F}_{im}^{coll}\in \mathbb{R}^{N_v \times {W} \times C}$.
Taking the collapsed images as value and key, and the 3D proposal features $F_{3d}$ originated from the LiDAR branch as query, we generate the query image features $\mathit{F}_{ca}$ by performing the cross-attention to achieve the feature interaction between the 3D proposal features $F_{3d}$ and the multi-view collapsed image features $\mathit{F}_{im}$. 
 \begin{equation}
    \mathit{F}_{ca} = \operatorname{CrossAtt}(\mathit{F}_{3d}, \mathit{F}_{im}^{coll}) \in \mathbb{R}^{N_{3d} \times C}.  
\end{equation}
Where $\operatorname{CrossAtt}$ is the cross-attention operation.
We further encode the center position of each 3D object proposal to a high-dimension 3D positional embedding $F_{3d}^{pos}$ to provide the 3D position knowledge. Then, we combine the features $F_{ca}$, $F_{3d}$, and $F_{3d}^{pos}$ in a concatenation manner and aggregate a compact feature representation by a multi-layer perception~(MLP). The aggregated feature is fed into an ($N_v+1$)-Category classification network to predict the view-level matching classification. 
\begin{equation}
    P_{cls} = \operatorname{MLP}(\operatorname{concat}(\mathit{F}_{ca}, \mathit{F}_{3d}, F_{3d}^{pos})) .
    \label{eq:vie_matching}
\end{equation} 
Where $P_{cls} \in \mathbb{R}^{N_{3d} \times (N_v+1)}$ denotes the output of view classification. We force the network to learn the classification given the ground truth pairs $V_{g}$ of 3D proposals and 2D views as supervision.
Note that we preserve the top-$K$ score indices of the targeted view and set $K$ = 2 as default, considering that a 3D object proposal might be detected in two views at most.

\subsection{Proposal-level Matching}
\label{subsec:proposallevel}

\begin{figure}[t!]
	\centering
	\includegraphics[width=0.99\linewidth]{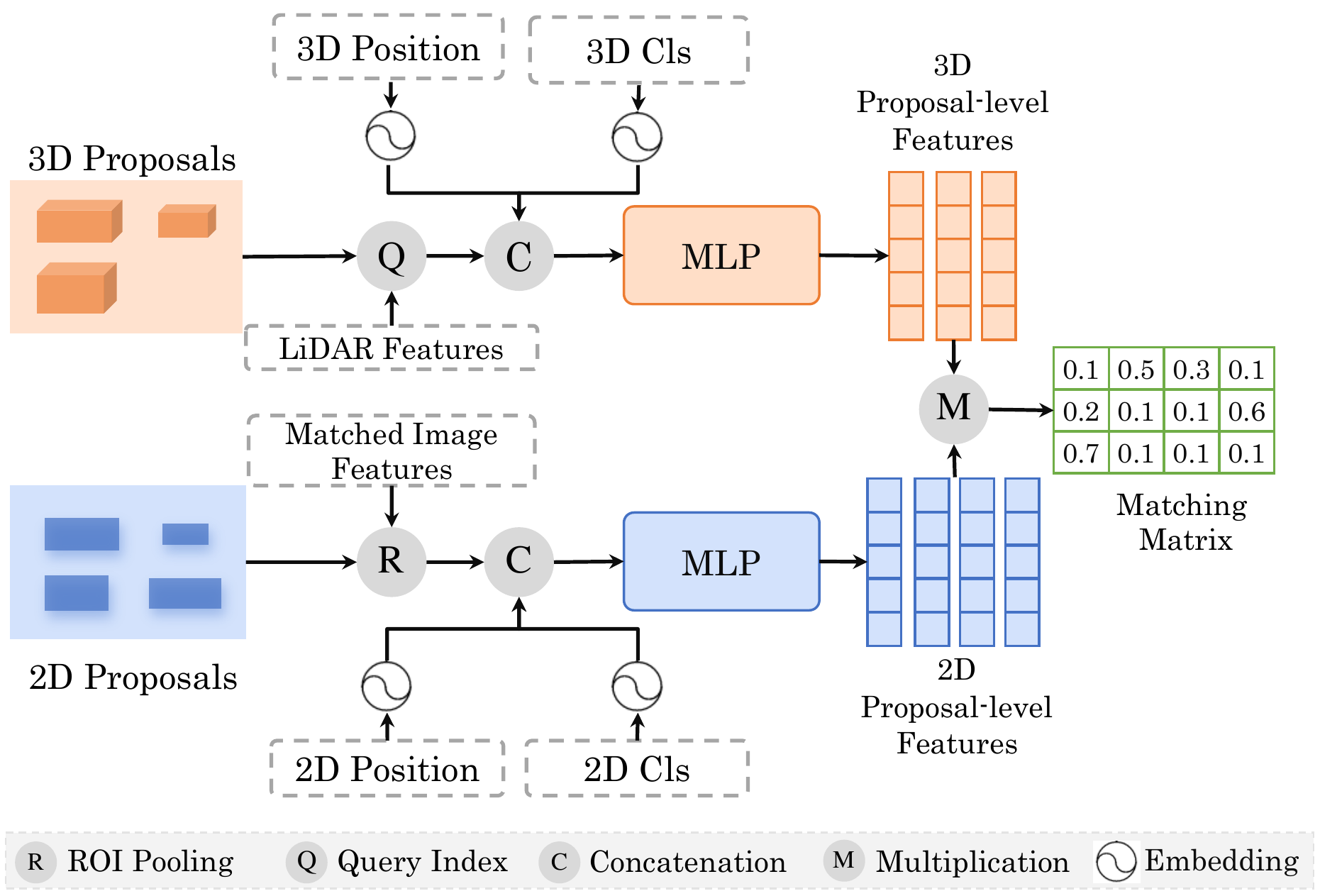}
	\caption{
The detail of Proposal-level Matching. Given the 2D proposals in the matched image view and the 3D proposals from LiDAR branch, we first independently extract the corresponding ROI features and concatenate them with their positional and classification embedding before feeding them into respective MLPs for computing the matching matrix.
		}
  \label{fig:proposal_level}
  \vspace{-8pt}
\end{figure}

After the view-level matching, we are able to associate the 3D object proposal with the most-relevant image-view features. Then the proposal-level matching is further applied to establish the proposal-wise correspondence between the 3D object proposals and 2D object proposals in the matched image view. 

Here, we denote the selected number of 3D proposals through view-level matching as $\hat{N}_{3d}$ and the number of 2D proposals in the corresponding view image as $\hat{N}_{2d}$.
The proposal-level matching is based on the similarity measure of features between two modalities, as shown in Figure~\ref{fig:proposal_level}. 
We obtain the corresponding matched image features $\hat{F}_{im} \subseteq F_{im}$ from the matched indices based on the formula~\ref{eq:vie_matching}.
We first implement a 2D ROI Pooling operation~\cite{girshick2015fast} to obtain 2D ROI features given each 2D detected bounding box and further aggregate the extracted 2D ROI features by an MLP to obtain a compact feature representation  $\hat{F}_{2d}^{roi} \in \mathbb{R}^{\hat{N}_{2d} \times C}$. We combine the $\hat{F}_{2d}^{roi}$ with 2D position embedding features $\hat{F}_{2d}^{pos} \in \mathbb{R}^{\hat{N}_{2d} \times C}$ by encoding the image coordinates of the predicted 2D proposals, as well as the 2D class-aware features $\hat{F}_{2d}^{cls} \in \mathbb{R}^{\hat{N}_{2d} \times C}$ by adopting an MLP to encode the predicted categories. After feeding the concatenation results into an MLP, we obtain the combined 2D proposal-level features as $\hat{F}_{2d}^{com} \in \mathbb{R}^{\hat{N}_{2d} \times C}$:
\begin{equation}
    \hat{F}_{2d}^{com} =\operatorname{MLP}(\operatorname{concat}(\hat{F}_{2d}^{roi},  \hat{F}_{2d}^{cls}, \hat{F}_{2d}^{pos} )).
\end{equation} 
Correspondingly, we then select the matched subset of the 3D proposal features from the LiDAR branch derived from the view-level matching stage and denote them as $\hat{F}_{3d} \in \mathbb{R}^{\hat{N}_{3d} \times C}$, such that $\hat{F}_{3d} \subseteq {F}_{3d}$. Similar with $\hat{F}_{2d}^{pos}$ and $\hat{F}_{2d}^{cls}$, we encode the 3D position and the category information of 3D proposal as $\hat{F}_{3d}^{pos}$ and $\hat{F}_{3d}^{cls}$. Then we aggregate them to obtain the combined 3D proposal-level features $\hat{F}_{3d}^{com} \in \mathbb{R}^{\hat{N}_{3d} \times C}$ by an MLP layer, which is formulated as:
\begin{equation}
    \hat{F}_{3d}^{com} =\operatorname{MLP}(\operatorname{concat}(\hat{F}_{3d},  \hat{F}_{3d}^{cls}, \hat{F}_{3d}^{pos} )) .
\end{equation} 
Considering that a 3D proposal may not have a matched 2D proposal, we append a full-zero 2D proposal feature to $\hat{F}_{2d}^{com}$ to indicate the unmatched 2D proposal and obtain the appended 2D proposal-level features $\overline{F}_{2d}^{com} \in \mathbb{R}^{(\hat{N}_{2d}+1)  \times C}$. Therefore, the final proposal-level matching matrix ${M}_p \in \mathbb{R}^{\hat{N}_{3d} \times (\hat{N}_{2d}+1)}$ can be computed as follows:
\begin{equation}
{M}_p = \frac{\hat{F}_{3d}^{com} \cdot (\overline{F}_{2d}^{com})^T}{\sqrt{C}} .
\label{eq:Mp}
\end{equation}
During the training stage, we add explicit supervision to guide the network to learn the matching matrix.
Assume $i$ and $j$ as the indices of the ordered 3D and 2D proposals, respectively, and $M_g$ represents the ground truth matrix, then we generate the pairs as:
\begin{equation}
{M_{g}}^{(i,j)}=\begin{cases}
1 &  i~\text{and}~j~\text{denote the same object,}\\
0 & \text{otherwise.}
\end{cases}
\end{equation}
We adopt cross-entropy loss for supervising the learning of the proposal-level matching matrix. During inference, there is no need for supervision. After obtaining the matching matrix $M_p$, we get the matched pairs between 2D and 3D proposals based on a fixed matching score threshold~(the default value is 0.1).




\subsection{Matching-based Fusion Module}
\label{subsec:fusion}

Based on the matching matrix $M_p$, we can obtain 2D ROI feature $\overline{F}_{2d}^{roi} \in \mathbb{R}^{\hat{N}_{3d}\times C}$ that matches one-to-one with $\hat{F}_{3d}$~(note that we set the unmatched 2D ROI features as zeros). 
Next, we enhance the LiDAR features absorbed from  the image features with rich semantic knowledge in the hierarchical three-pronged strategy to perform the fusion.

\noindent\textbf{Query-Pixel Fusion.} We adopt the transformer decoder layer following the design of DETR\cite{carion2020end} to fuse the LiDAR queries and the pixel-wise features within the 2D RoI region, which includes a self-attention operation and a cross-attention operation. For simplicity, we abbreviate this operation as $\operatorname{Decoder}(q,k,v,m)$, where $q$, $k$, $v$ and $m$ mean the query, key, value and mask. The self-attention between LiDAR object queries learns the relationship between different object candidates and the cross attention between LiDAR queries $\hat{F}_{3d}$~(as $q$) and the matched image feature maps $\hat{F}_{im}$ ~(as $k$ and $v$) aggregates semantic context onto the object candidates. It can be formulated as:
\begin{equation}
O_1 = \operatorname{Decoder}(\hat{F}_{3d}, \hat{F}_{im}, \hat{F}_{im}, \hat{M}_{2d}^{roi}). 
\label{eq:query_o1}
\end{equation}
Where $\hat{M}_{2d}^{roi}$ denotes the mask that we only retain the pixel-wise features of the matched image within the 2D ROI regions. The details  of $\operatorname{Decoder}$ are presented in the supplementary material.

\noindent\textbf{Query-}$\mathbf{{{ROI}_{2d}}}$ \textbf{Fusion.} We also implement the ROI-level fusion with the LiDAR queries by simply concatenating the LiDAR queries and the weighted 2D RoI image features and adopting an MLP for feature fusion.
\begin{equation}
O_2 = \mathrm{MLP}(\mathrm{Concat}[\hat{F}_{3d}, S\odot \overline{F}_{2d}^{roi}]).
\label{eq:query_o2}
\end{equation}
Where $S \in \mathbb{R}^{\hat{N}_{3d}\times 1}$ is the maximum matching score computed by the matching matrix in formula~\ref{eq:Mp} along each row and is applied to reweight the 2D ROI features. $\odot$ means element-wise multiplication with the broadcast operation.

\noindent\textbf{$\mathbf{ROI_{3d}}$
- $\mathbf{ROI_{2d}}$ 
Fusion.} We can also perform ROI-level 2D -3D feature fusion by decoding the LiDAR ROI features $\hat{F}_{3d}^{roi}$ by 3D ROI Pooling~\cite{shi2019pointrcnn} and the image ROI features $\hat{F}_{2d}^{roi}$ using the similar decoder layer in formula~\ref{eq:query_o1}.
\begin{equation}
O_3 = \operatorname{Decoder}(\hat{F}_{3d}^{roi}, 
\overline{F}_{2d}^{roi}, \overline{F}_{2d}^{roi}).  
\label{eq:query_o3}
\end{equation}
Finally, we further fuse these three-level features in a concatenation manner and feed it to a feed-forward network~(FFN) to produce the final predictions.
\begin{equation}
\operatorname{Pred} = \operatorname{FFN}(\operatorname{concat}(O_1, O_2, O_3)). 
\label{eq:querycomb}
\end{equation}

\subsection{Loss Function}
\label{subsec:trainingloss}



For simplicity, we define the 3D detection task loss $L_{det}$ as in ~\cite{bai2022transfusion}, which includes the regression loss of 3D bounding boxes and the classification loss for the categories. 
Furthermore, we add two additional losses for the two-level matching, \textit{i.e.}, the view-level matching loss and the proposal-level matching loss denote as $L_{view}$ and $L_{pro}$. The total loss $L_{total}$ is formulated as:
\begin{equation}
 L_{total}  = L_{det} + \lambda_1 L_{view} + \lambda_2 L_{pro},
\end{equation}
where $\lambda_1$ and $\lambda_2$ represent the balanced weights for the view-level matching loss and the proposal-level matching loss, respectively. $L_{view}$ and $L_{pro}$ are both cross-entropy losses (denoted by $\mathrm{CE}$), \ie, $L_{view}$ = $\mathrm{CE}(P_{cls}, V_{g})$, and $L_{pro}$ = $\mathrm{CE}({M}_p, M_{g})$.

\section{Experiments}
\label{sec:exp}

\begin{table*}[t]
\small
\setlength{\tabcolsep}{15pt}
\resizebox{1.0\linewidth}{!}{
\centering
\begin{tabular}{l|c|c|c||c|c|c}
\hline
        \multirow{2}{*}{Method}  & \multicolumn{1}{c|}{\multirow{2}{*}{Backbone}}   & \multicolumn{2}{c||}{Input} & \multicolumn{3}{c}{Disturbed Input}\\
        \cline{3-7} 
        & \multicolumn{1}{c|}{} & \multicolumn{1}{c|}{L} & \multicolumn{1}{c||}{LC} & \multicolumn{1}{c|}{LC+AS} & \multicolumn{1}{c|}{LC+MSP} & \multicolumn{1}{c}{LC+DI}\\
         \hline
        MVP~\cite{yin2021multimodal} &Pillar-based &52.40 &62.72   &20.88~(\robcolor{-41.84})  & 32.77(\robcolor{-29.95}) &52.20~(\robcolor{-10.52}) \\
        TransFusion~\cite{bai2022transfusion} &Pillar-based &55.04 &59.96  &51.36~(\robcolor{-8.60}) &52.39~(\robcolor{-7.57}) &51.38~(\robcolor{-8.58}) \\
        \mycolor{\ourMethod{}} &\mycolor{Pillar-based}   &\mycolor{\textbf{55.04}} &\mycolor{\textbf{59.88}}  &\mycolor{\textbf{58.48}}~(\robcolor{\textbf{-1.40}}) &\mycolor{\textbf{58.80}}~(\robcolor{\textbf{-1.08}}) &\mycolor{\textbf{56.96}}~(\robcolor{\textbf{-2.92}}) \\
        \hline
        MVP~\cite{yin2021multimodal} &Voxel-based &59.53 &65.97 &28.54~(\robcolor{-37.43}) &35.86~(\robcolor{-30.11})   &57.67~(\robcolor{-8.30}) \\
        TransFusion~\cite{bai2022transfusion} &Voxel-based &64.98 &67.06 &65.54~(\robcolor{-1.52}) &65.34~(\robcolor{-1.72})  &64.42~(\robcolor{-2.64})   \\
        BEVFusion~\cite{liu2022bevfusion} &Voxel-based  &64.68 &68.52&64.19~(\robcolor{-4.33})  &63.41~(\robcolor{-5.11})  &64.64~(\robcolor{-3.88}) \\
        \mycolor{\ourMethod{}} &\mycolor{Voxel-based}  &\mycolor{\textbf{64.98}} &\mycolor{\textbf{66.93}} &\mycolor{\textbf{66.66}}~(\robcolor{\textbf{-0.27}}) &\mycolor{\textbf{65.88}}~(\robcolor{\textbf{-1.05}})   &\mycolor{\textbf{65.59}}~(\robcolor{\textbf{-1.34}}) \\
        \hline
	\end{tabular}
 }
	\caption{Robustness for existing methods on nuScenes validation set. `Pillar-based' and `Voxel-based'  stand for the  backbones of PointPillars~\cite{lang2019pointpillars} and VoxelNet~\cite{zhou2018voxelnet}. `L' and `LC' indicate the LiDAR-only and LiDAR-Camera fusion methods. AS means Asynchronous Sensors~(0.5s), MSP means Misaligned Sensor Placement ~(Medium level), and DI means Degenerated Images~(3 dropped images). We present the dropped performance of fusion methods after introducing AS, MSP and DI in \textbf{\robcolor{blue}}. We evaluate the results with mAP. } 
 \label{tab:exp_total_robust}
\end{table*}

\begin{table*}[t]
\small
\setlength{\tabcolsep}{10pt}
\resizebox{1.0\linewidth}{!}{
\centering
\begin{tabular}{l|c|c|c|c|c|c}
\hline
        \multirow{2}{*}{Method}  & \multicolumn{1}{c|}{\multirow{2}{*}{Backbone}}   & \multicolumn{5}{c}{Asynchronous Times~(s)}  \\
        \cline{3-7} 
        & \multicolumn{1}{c|}{} & \multicolumn{1}{c|}{0.08} & \multicolumn{1}{c|}{0.25} & \multicolumn{1}{c|}{0.50} & \multicolumn{1}{c|}{1.00} & \multicolumn{1}{c}{2.00}\\
         \hline
        MVP~\cite{yin2021multimodal} &Pillar-based &41.64~(\robcolor{-21.08}) &35.54~(\robcolor{-27.18})  &20.88~(\robcolor{-41.84}) &13.31~(\robcolor{-49.41})  &9.20~(\robcolor{-53.52}) \\
        TransFusion~\cite{bai2022transfusion} &Pillar-based &56.02~(\robcolor{-3.94}) &53.10~(\robcolor{-6.86}) &51.36~(\robcolor{-8.60}) &48.67~(\robcolor{-11.29}) &46.79~(\robcolor{-13.17}) \\
        \mycolor{\ourMethod{}} &\mycolor{Pillar-based}  &\mycolor{\textbf{59.69~(\robcolor{-0.19})}} &\mycolor{\textbf{59.26~(\robcolor{-0.62})}} &\mycolor{\textbf{58.80}}~(\robcolor{\textbf{-1.08}}) &\mycolor{\textbf{58.11}}~(\robcolor{\textbf{-1.77}}) &\mycolor{\textbf{57.38}}~(\robcolor{\textbf{-2.50}}) \\
        \mycolor{Over \ourMethod{}-L} &\mycolor{Pillar-based} &\mycolor{\textbf{\redcolor{+4.65}}} &\mycolor{\textbf{\redcolor{+4.22}}} &\mycolor{\textbf{\redcolor{+3.76}}}  &\mycolor{\textbf{\redcolor{+3.07}}} &\mycolor{\textbf{\redcolor{+2.34}}} \\
        \hline
        MVP~\cite{yin2021multimodal} &Voxel-based &50.88~(\robcolor{-15.09}) &47.37~(\robcolor{-18.60})  &28.54~(\robcolor{-37.43}) &21.86~(\robcolor{-44.11})  &17.76~(\robcolor{-48.21}) \\
        TransFusion~\cite{bai2022transfusion} &Voxel-based &66.51~(\robcolor{-0.55}) &65.88~(\robcolor{-1.18}) &65.54~(\robcolor{-1.52}) &65.11~(\robcolor{-1.95}) &64.61~(\robcolor{-2.45})   \\
        BEVFusion~\cite{liu2022bevfusion} &Voxel-based  &66.57~(\robcolor{-1.95}) &65.11~(\robcolor{-3.41}) &64.19~(\robcolor{-4.33}) &62.89~(\robcolor{-5.63}) &61.81~(\robcolor{-6.71}) \\
        \mycolor{\ourMethod{}} &\mycolor{Voxel-based}  &\mycolor{\textbf{66.92~(\robcolor{-0.01})}} &\mycolor{\textbf{66.80~(\robcolor{-0.13})}} &\mycolor{\textbf{66.66}}~(\robcolor{\textbf{-0.27}}) &\mycolor{\textbf{66.27}}~(\robcolor{\textbf{-0.66}})  &\mycolor{\textbf{65.93}}~(\robcolor{\textbf{-1.00}}) \\
        \mycolor{Over \ourMethod{}-L} &\mycolor{Voxel-based} &\mycolor{\textbf{\redcolor{+1.94}}} &\mycolor{\textbf{\redcolor{+1.82}}} &\mycolor{\textbf{\redcolor{+1.68}}} &\mycolor{\textbf{\redcolor{+1.29}}} &\mycolor{\textbf{\redcolor{+0.95}}}  \\
        \hline
	\end{tabular}
 }
	\caption{Robustness for asynchronous sensors on nuScenes. `*-L' means the  LiDAR-only detector of *. We evaluate the results with mAP.} 
 \label{tab:robust_asyn}
 \vspace{-8pt}
\end{table*}

\subsection{Datasets and Evaluation Metrics}

To verify the effectiveness of our method, we conduct extensive experiments on nuScenes dataset~\cite{caesar2020nuscenes}, which is a large-scale multi-modal benchmark collected by vehicles equipped with one LiDAR, six cameras, and five Radars. Each scene is composed of 20\textit{s} video frames with 3D bounding boxes every 0.5\textit{s} fully annotated. We follow the official splits and divide 1000 scenes into 700, 150, and 150 scenes as the \textit{train}, \textit{validation} and \textit{test} set. The nuScenes detection score~(NDS) and the mean average presion~(mAP) for 10 foreground classes are adopted as the evaluation metrics for the 3D detection task.

\subsection{Implementation Details}
\noindent\textbf{{Network Setting.}} 
The whole framework includes an image branch and a LiDAR branch. The image branch adopts the popular 2D detector Mask-RCNN~\cite{he2017mask} based on ResNet-50~\cite{he2016deep}. The input camera images are downsampled to 448 $\times$ 800, which is 1/2 size of the original resolution 896 $\times$ 1600 to save the computation cost. Besides, FPN~\cite{lin2017feature} is applied to fuse multi-scale camera features to produce robust feature maps. For the LiDAR branch, we adopt TransFusion-L based 
 on both VoxelNet~\cite{zhou2018voxelnet} and PointPillars~\cite{lang2019pointpillars} backbones. The detection range  is set to [-54, 54], [-54, 54], [-5, 3] meters along the $X$, $Y$, and $Z$ axes, respectively. 
We set the pillar size of [0.2, 0.2, 8] meters  for the Pillar-based  backbone, and the voxel size of [0.075, 0.075, 0.2] meters for the Voxel-based backbone. In the training stage, we adopt 200 queries for a fair comparison with Transfusion~\cite{bai2022transfusion}. If not specified, we use 200 queries as the default setting in the inference stage.

\noindent\textbf{Training and Inference Schemes.} We keep exactly the same training scheme and data augmentation as TransFusion~\cite{bai2022transfusion} for a fair comparison. The camera branch  is first pre-trained on COCO~\cite{lin2014microsoft} and then further trained on nuImage~\cite{caesar2020nuscenes} following TransFusion~\cite{bai2022transfusion} and the weights are kept frozen during training. 
The LiDAR branch is trained for 20 epochs and the fusion module for another 6 epochs. To optimize our network, we apply AdamW~\cite{kingma2014adam} optimizer with a max learning rate of $1 \times 10^{-4}$ and weight decay of 0.01. In the total loss function, we set the  balanced weights of $\lambda_1$ and $\lambda_2$ as 0.2 and 0.1 by default. Note that the nuScenes dataset does not directly provide the ground truth of the view-matching matrix $V_{g}$ and $M_g$, thus, we produce the pairs of two modalities by the projection matrix only in the training process. In the inference stage, we discard the projection matrix by learning box matching.




\subsection{Calibration-free Framework for Inference}
In this subsection, we verify the robustness of our calibration-free framework for multi-modal fusion in challenging cases including asynchronous sensors, misaligned sensor placement as well as degenerated images on nuScenes \emph{validation} set. For comparison, we select three representative projection-based fusion approaches, including MVP~\cite{yin2021multimodal}, TransFusion~\cite{bai2022transfusion} and  BEVFusion~\cite{liu2022bevfusion}.
Before verifying the robustness, we first present the summarized results of the above methods in Table~\ref{tab:exp_total_robust} under the normal cases and the challenging cases with disturbed inputs. 
As shown, under normal circumstances, when combining the image features and LiDAR features, MVP~\cite{yin2021multimodal}, TransFusion~\cite{bai2022transfusion} and BEVFusion~\cite{liu2022bevfusion} obtain measurable gains over the corresponding LiDAR-only counterparts. However, when adding disturbed inputs, all three projection-based methods suffer from heavier performance degradation than our \ourMethod{}, which illustrates the superiority of our matching-based fusion mechanism. In the following, we will introduce the details of the disturbed inputs and the performance comparisons under these challenging cases.

\noindent\textbf{Robustness to Asynchronous Sensors.} Asynchronous issue is a common and challenging factor that may damage the efficacy of existing projection-based methods.  
To investigate the impacts, we simulate this case by making the two modalities asynchronous in the temporal dimension. 
Specifically, we take the LiDAR's moment as the dominant and then pick the image frame advancing it by $t$ image frames~(each frame interval is about 0.08 seconds since the cameras run at 12Hz on nuScenes). We conduct experiments with asynchronous times of $0.08$\textit{s}, $0.25$\textit{s}, $0.50$\textit{s}, $1.00$\textit{s} and $2.00$\textit{s}, whose results are summarized in Table~\ref{tab:robust_asyn}. 
First, we can observe that projection-based methods including MVP~\cite{yin2021multimodal}, TransFusion~\cite{bai2022transfusion} and  BEVFusion~\cite{liu2022bevfusion} are more likely to be affected by asynchronous issues. As the asynchronous interval increases, the performance reduction is more and more obvious. However, our method demonstrates strong robustness to this issue. 
Besides, note that our \ourMethod{}  still produces satisfactory performance by outperforming the corresponding LiDAR-only baseline in different backbones even when the asynchronous time reaches 2.00 seconds. The main reason is that our matching-based fusion can pick out the best-matched ROI image feature to enhance the corresponding LiDAR feature if matched~(\textit{e.g.}, the matching  score is greater than the threshold of 0.1). Meanwhile, if not matched, \ourMethod{} can disable the process of the LiDAR feature enhanced by the ROI image features, thus leaving the LiDAR feature undisturbed. 
 

\begin{table}[t]
\small
\setlength{\tabcolsep}{2pt}
\resizebox{1.0\linewidth}{!}{
\centering
\begin{tabular}{l|c|c|c|c}
\hline
        \multirow{2}{*}{Method}  & \multicolumn{1}{c|}{\multirow{2}{*}{BK}}   & \multicolumn{3}{c}{Misaligned Sensor Placement}  \\
        \cline{3-5} 
        & \multicolumn{1}{c|}{} & \multicolumn{1}{c|}{Small} & \multicolumn{1}{c|}{Medium} & \multicolumn{1}{c}{Large} \\
         \hline
        MVP~\cite{yin2021multimodal} &P &48.57~(\robcolor{-14.15}) &32.77~(\robcolor{-29.95})  &21.68~(\robcolor{-41.04}) \\
        TransFusion~\cite{bai2022transfusion} &P &57.39~(\robcolor{-2.57}) &52.39~(\robcolor{-7.57}) &49.44~(\robcolor{-10.52}) \\
        \mycolor{\ourMethod{}} &\mycolor{P}  &\mycolor{\textbf{59.49}}~(\robcolor{\textbf{-0.39}}) &\mycolor{\textbf{58.48}}~(\robcolor{\textbf{-1.40}}) &\mycolor{\textbf{57.65}}~(\robcolor{\textbf{-2.23}}) \\
        \mycolor{Over \ourMethod{}-L} &\mycolor{P} &\mycolor{\textbf{\redcolor{+4.45}}} &\mycolor{\textbf{\redcolor{+3.44}}}&\mycolor{\textbf{\redcolor{+2.61}}} \\
        \hline
        \hline
        MVP~\cite{yin2021multimodal} &V  &51.24~(\robcolor{-14.73}) &35.86~(\robcolor{-30.11})  &25.37~(\robcolor{-40.60}) \\
        TransFusion~\cite{bai2022transfusion} &V  &66.35~(\robcolor{-0.71}) &65.34~(\robcolor{-1.72}) &64.45~(\robcolor{-2.61})   \\
        BEVFusion~\cite{liu2022bevfusion} &V   &66.32~(\robcolor{-2.20}) &63.41~(\robcolor{5.11}) &62.02~(\robcolor{-6.50}) \\
        \mycolor{\ourMethod{}} &\mycolor{V}  &\mycolor{\textbf{66.51}}~(\robcolor{\textbf{-0.42}}) &\mycolor{\textbf{65.88}}~(\robcolor{\textbf{-1.05}})  &\mycolor{\textbf{65.16}}~(\robcolor{\textbf{-1.77}}) \\
        \mycolor{Over \ourMethod{}-L} &\mycolor{V} &\mycolor{\textbf{\redcolor{+1.53}}} &\mycolor{\textbf{\redcolor{+0.90}}} &\mycolor{\textbf{\redcolor{+0.18}}} \\
        \hline
	\end{tabular}
 }
	\caption{Robustness for misaligned sensor placement on nuScenes \emph{validation} set. The misaligned placement of `Small', `Medium' and `Large' means that LiDAR points are rotated 1.5, 3.0, 5.0 degrees along the vertical direction and are translated 0.15, 0.30 and 0.50 meters, respectively. `P' and `V' are short for Pillar-based and Voxel-based. We evaluate the results with mAP.} 
 \label{tab:robust_placement} 
 \vspace{-8pt}
\end{table}

\noindent\textbf{Robustness to Misaligned Sensor Placement.} 
In practice, it is more likely to have sensors offline-calibrated for data collection and train the multi-modal fusion methods and deploy the model for online inference. However, there may still exist hidden sensor displacements even for the same self-driving vehicle because of bad road conditions, causing the distribution discrepancy between train data and test data. We mimic this real situation by perturbing the input LiDAR points on test data but reserving the original calibration parameters. For simplicity, we perturb LiDAR points by rotating and translating them to varying degrees to simulate the small, medium, and large misaligned sensor placement, whose results are provided in Table~\ref{tab:robust_placement}.
We observe that voxel-based fusion methods are generally more robust than pillar-based manners under the same case.
The reason may be that the weaker pillar-based approaches rely more on utilizing rich image semantic features to enhance LiDAR features for better performance, which is more harmful when the fused image features are unreliable. 
Especially, the performance of TransFusion~\cite{bai2022transfusion} and MVP~\cite{yin2021multimodal} with pillar-based backbones are much lower than their LiDAR-only counterparts under the large misaligned sensor placement, respectively.  
These indicate the criticality of accurate sensor placement for existing projection-based methods.
However, \ourMethod{} can still establish the box-level correspondence between two modalities benefiting from  box matching and thus maintain steady performance. Finally, in the supplementary materials, we also provide the results in the case of the poor projection matrix due to inaccurate calibrations, in which the performance of \ourMethod{} will not be declined thanks to our calibration-free framework.

\begin{table}[t]
\small
\setlength{\tabcolsep}{3pt}
\resizebox{1.0\linewidth}{!}{
\centering
\begin{tabular}{l|c|c|c|c}
\hline
        \multirow{2}{*}{Method}  & \multicolumn{1}{c|}{\multirow{2}{*}{BK}}   & \multicolumn{3}{c}{Dropped Images}  \\
        \cline{3-5} 
        & \multicolumn{1}{c|}{} & \multicolumn{1}{c|}{1} & \multicolumn{1}{c|}{3} & \multicolumn{1}{c}{6} \\
         \hline
        MVP~\cite{yin2021multimodal} &P &61.51~(\robcolor{-1.21}) &52.20~(\robcolor{-10.52})  &16.61~(\robcolor{-46.11}) \\
        TransFusion~\cite{bai2022transfusion} &P &56.14~(\robcolor{-3.82}) &51.38~(\robcolor{-8.58}) &44.64~(\robcolor{-15.32}) \\
        \mycolor{\ourMethod{}} &\mycolor{P}  &\mycolor{\textbf{58.96}}~(\robcolor{\textbf{-0.92}}) &\mycolor{\textbf{56.96}}~(\robcolor{\textbf{-2.92}})  &\mycolor{\textbf{54.42}}~(\robcolor{\textbf{-5.46}}) \\
        \mycolor{Over \ourMethod{}-L} &\mycolor{P} &\mycolor{\textbf{\redcolor{+3.93}}} &\mycolor{\textbf{\redcolor{+1.92}}} &\mycolor{\textbf{\redcolor{-0.62}}} \\
        \hline
        MVP~\cite{yin2021multimodal} &V  &65.34~(\robcolor{-0.63}) &57.67~(\robcolor{-8.30})  &24.71~(\robcolor{-41.26}) \\
       TransFusion~\cite{bai2022transfusion} &V  &65.89~(\robcolor{-1.17}) &64.42~(\robcolor{-2.64}) &63.56~(\robcolor{-3.50})  \\
        BEVFusion~\cite{liu2022bevfusion} &V   &67.29~(\robcolor{-1.23}) &64.64~(\robcolor{-3.88}) &61.24~(\robcolor{-7.28}) \\
        \mycolor{\ourMethod{}} &\mycolor{V}  &\mycolor{\textbf{66.57}}~(\robcolor{\textbf{-0.36}}) &\mycolor{\textbf{65.59}}~(\robcolor{\textbf{-1.34}})  &\mycolor{\textbf{64.78}}~(\robcolor{\textbf{-2.15}}) \\
        \mycolor{Over \ourMethod{}-L} &\mycolor{V} &\mycolor{\textbf{\redcolor{+1.59}}} &\mycolor{\textbf{\redcolor{+0.61}}} &\mycolor{\textbf{\redcolor{-0.20}}} \\
        \hline
	\end{tabular}
 }
	\caption{Robustness for degenerated camera images on nuScenes \emph{validation} set. `P' and `V' are short for Pillar-based and Voxel-based, respectively. We evaluate the results with mAP. }
	\label{tab:robust_drop}
 \vspace{-8pt}
\end{table}

\noindent\textbf{Robustness to Degenerated Camera Images.} To verify the robustness to the extreme condition of dropped images, we follow TransFusion~\cite{bai2022transfusion} and randomly discard one or more images in a frame. In Table~\ref{tab:robust_drop}, MVP~\cite{yin2021multimodal} degrades the performance dramatically as the number of discarded images increases due to the tightly-coupled manner. A similar conclusion can be drawn for BEVFusion~\cite{liu2022bevfusion}. Compared with them, TransFusion~\cite{bai2022transfusion} with the voxel-based backbone can keep a stable performance benefiting from the soft association fusion strategy, which can adaptively combine the related image features. However, we find this strategy does not work well for the weaker pillar-based backbone. We argue that the soft association strategy does not effectively decouple the two modalities.
Besides,  when dropping three images, the above methods even produce inferior performances against the corresponding LiDAR-only baselines. We attribute the phenomena to the fact that these methods tightly couple image and point cloud modalities, which will be greatly affected when the image modality is severely disturbed. 
On the contrary, our \ourMethod{} consistently exhibits a competitive performance regardless of the backbones thanks to the decoupling of two modalities by the proposed box-matching manner. Finally, for dropping six images, it may be necessary for these projection-based fusion methods to switch to LiDAR-only detectors if possible. However, under this case, our \ourMethod{} still maintains a comparable performance to the corresponding LiDAR-only detector, which illustrates its strong robustness to degenerated images.
For more discussions on whether to fall back on LiDAR-only detectors and more experiments about degenerated images~(\textit{e.g.}, adding noise to images), please refer to the supplemental materials.

\subsection{Ablation Studies}

In this part, we mainly conduct some relevant ablation studies on view-level matching and proposal-level matching on nuScenes \emph{validation} set. For more ablation studies~(\eg, different fusion mechanism in the formula~\ref{eq:querycomb}) about \ourMethod{}, please refer to our supplemental materials.

\begin{table}[h]
\small
\centering
\setlength{\tabcolsep}{16pt}
\resizebox{1.0\linewidth}{!}{
\begin{tabular}{l|c|cc}
\hline
         View-Level  & Acc. &mAP & NDS \\
         \hline
         Top-1  & 85.94 & 66.34 & 70.52  \\
         \mycolor{Top-2}  & \mycolor{\textbf{98.46}} & \mycolor{\textbf{66.93}} & \mycolor{\textbf{70.90}}  \\
        \hline
	\end{tabular}
 }
	\caption{Ablation study for view-level matching. `Acc.' indicates the classification accuracy for view matching.}
	\label{tab:ab_view}
 \vspace{-8pt}
\end{table}

\begin{table}[h]
\small
\centering
\setlength{\tabcolsep}{24pt}
\resizebox{1.0\linewidth}{!}{
\begin{tabular}{l|cc}
\hline
         Matching  &mAP & NDS \\
         \hline
         One Level  & 66.48 & 70.57  \\
         \mycolor{Two Level}  & \mycolor{\textbf{66.93}} & \mycolor{\textbf{70.90}}  \\
        \hline
	\end{tabular}
 }
	\caption{Ablation for our two-level matching design. One-level and Two-level denote adopting only proposal-level matching and combining view-level with proposal-level matching, respectively.}
	\label{tab:two_level_matching}
 \vspace{-5pt}
\end{table}

\noindent\textbf{Top-2 \emph{vs.} Top-1 View-level Matching.} In Table~\ref{tab:ab_view}, we present the results of selecting Top-1 and Top-2 view matching predictions to analyze their effects for final 3D detection performance. Correct view-matching is critical to promise the final object matching due to the view-level matching followed by the proposal-level matching. As we can observe in Table~\ref{tab:ab_view}, the Top-2 manner has a  superior accuracy of 98.46\% for view matching, which is much higher than that of Top-1.
Moreover, the choice of Top-2 view-level matching is in line with the fact that there is a certain overlapping area between the multiple views. Finally, the manner based on Top-2 surpasses the Top-1 with an mAP of 0.59\%, which indicates the reliability of selecting Top-2 view-level matching predictions for the subsequent proposal-level matching.

\noindent\textbf{Two-level Matching \emph{vs.} One-level Matching.} We further study the merits of two-level matching over one-level matching (\ie, only utilizing the proposal-level matching through computing the proposal-level matrix of all 3D proposals with 2D proposals in each image.). As shown in Table~\ref{tab:two_level_matching}, the manner of adopting two-level matching outperforms the one-level method with an mAP of 0.45\%. We assume that the first level (view-level matching) filters out the unmatched 3D proposals to specific camera views, which reduces the risk of false matches for feature enhancements to some extent. Furthermore, we also empirically observe that the model with one-level matching will take much more time to converge. These results prove the efficacy and efficiency of adopting two-level matching. 



\begin{table}[t]
\small
\setlength{\tabcolsep}{3pt}
\resizebox{1.0\linewidth}{!}{
	\centering
	\begin{tabular}{l|c|c|c|cc}
\hline
         {Method} & {Publishion}& {Input} &Proj.  & {NDS}~$\uparrow$ & {mAP}~$\uparrow$  \\
         \hline
        TransFusion-L~\cite{bai2022transfusion} &CVPR2022 &L &--  &70.0 &65.0  \\
         TransFusion~\cite{bai2022transfusion}$^\dag$ &CVPR2022 &LC  &\cmark   & 70.9 &67.5   \\ 
         \mycolor{\ourMethod{}~(Ours)}$^\dag$ &\mycolor{--} &\mycolor{LC} &\mycolor{\textbf{\xmark}}   & \mycolor{{71.1}} & \mycolor{{67.4}}  \\
         \hline
        \hline
        PointPillars~\cite{lang2019pointpillars} &CVPR2019 & L &--  & 45.3 & 30.5 \\
        {CenterPoint~\cite{yin2021center}}$^\ddagger$ &CVPR2021 & L &--   & 67.3 & 60.3  \\
         {TransFusion-L~\cite{bai2022transfusion}} &CVPR2022 & L &-- & {70.2}  & {65.5} \\
          {PointPainting\cite{vora2020pointpainting}} &CVPR2020 & LC  &\cmark   & 58.1 & 46.4  \\
          {3D-CVF~\cite{yoo20203d}} &ECCV2020 & LC &\cmark     & 62.3 & 52.7  \\
          {PointAugmenting~\cite{wang2021pointaugmenting}$^\ddagger$} &CVPR2021 & LC &\cmark   & 71.0 & 66.8  \\
          {MVP~\cite{yin2021multimodal}} &NeurIPS2021 & LC &\cmark    & 70.5  & 66.4 \\
          Focals Conv-F~\cite{chen2022focal} &CVPR2022 &LC & \cmark &71.8 &67.8 \\
          VirConv~\cite{liang2022bevfusion} &CVPR2023 & LC & \cmark   & 72.3 & 68.7  \\
         TransFusion~\cite{bai2022transfusion} &CVPR2022 & LC &\cmark   &71.7  & 68.9  \\
         \mycolor{\ourMethod{}~(Ours)}  &\mycolor{--} & \mycolor{LC} & \mycolor{\textbf{\xmark }}  & \mycolor{{72.1}}  & \mycolor{{68.9}} \\ 
        \hline
	\end{tabular}
 }
	\caption{Results on the nuScenes \emph{validation} (top) and \emph{test} (bottom) set. `Proj.' indicates that using the calibration for projection-based fusion. `L' and `C' represent LiDAR and Camera, respectively. $^\dag$ represents the results with more queries~(\textit{e.g.}, 500) in the inference stage. $^\ddagger$ means adopting Test-Time Augmentation~(TTA). Note that we do not adopt any TTA or multi-model ensemble strategy during the inference stage for both the \emph{validation} and \emph{test} set.
}
\label{tab:nuscene_test}
\vspace{-8pt}
\end{table}

\subsection{Comparison with State-of-the-arts}
Table~\ref{tab:nuscene_test} presents the comparison with representative methods on the nuScenes \emph{validation}~(top) and \emph{test}~(bottom) splits. 
On the \emph{validation} set, our method obtains comparable performance to TransFusion~\cite{bai2022transfusion}.
On the \emph{test} split, compared with our baseline model TransFusion-L~\cite{bai2022transfusion}, 
\ourMethod{} brings a gain with 3.4\% mAP and 1.9\% NDS, which even achieves higher NDS over the Transfusion~\cite{bai2022transfusion}. Meanwhile, our method obtains competitively close performance 
to the advanced fusion method  VirConv~\cite{VirConv} on \emph{test} set. This illustrates the effectiveness of our matching-based fusion. 
Besides, it is noteworthy that, unlike previous multi-modal fusion methods, the main goal of this work is not about designing a sophisticated fusion model but mitigating one critical and easily overlooked issue, which is freeing up the heavy dependence on an accurate projection matrix to achieve robust detection during inference. And the proposed box-matching solution proves its great potential to handle these difficult scenarios for autonomous vehicles.
Finally, we provide the qualitative visualization of  \ourMethod{} and the learned matching in supplemental materials.

\subsection{Limitations}
\label{sec:limitions}

Our method utilizes 3D proposals from LiDAR as the query for associations with 2D proposals, which is based on the assumption that LiDARs are more powerful perception sensors than cameras. However, for small and distant objects, the LiDAR may fail to detect them, \ie~, if no corresponding 3D proposals are generated, then our method may face challenges in recovering them. By contrast, camera images can better deal with these issues. In the future, we will further explore ways on improving the fusion process by utilizing the camera information.

\section{Conclusion} 
In this paper, we have pointed out the common and crucial challenging cases for existing multi-modal fusion methods in real-world self-driving systems, such as  asynchronous sensors, misaligned sensor placement and degenerated camera images. To address these issues, we have proposed a novel multi-modal fusion network name \ourMethod{} by the mechanism of box matching. In extensive experiments, the effectiveness and robustness of our \ourMethod{} have been verified in these challenging cases, which may provide some new insights for the next multi-modal 3D detection. In the future, we will further explore the robustness for more extreme conditions and apply our FBMNet to more advanced 3D detection frameworks.

{\small
\bibliographystyle{ieee_fullname}
\bibliography{egbib}
}

\clearpage
\appendix
\section{Appendix}

The supplementary materials are organized as follows. First, we describe the details of the Decoder in the matching-based fusion module in section \ref{sec:decoder}. Then, we present more experiments on robustness in section ~\ref{sec:robust_exp}, which includes the settings of inaccurate calibration matrix, noisy images and multiple disturbances.  Next, we provide more ablation studies in section~\ref{sec:more_abl}, which involves the comparison of matching-based fusion and global cross-attention fusion, and the ablation about fusion manners. Besides, we show the qualitative visualization of \ourMethod{} and the learned matching results on the nuScenes validation dataset in section \ref{sec:vis}. Finally, we discuss some limitations of our \ourMethod{} in section \ref{sec:limitions}. \textit{Code will be released.}

\label{sec:decoder}

\section{Details of {Decoder} in Matching-based Fusion Module}
\label{sec:decoder}

In the matching-based fusion module, we adopt the transformer decoder following the design of DETR~\cite{carion2020end}. Specifically, the decoder includes a self-attention operation and a cross-attention operation. For simplicity, we abbreviate this operation as $\operatorname{Decoder}(q,k,v,m)$, where $q$, $k$, $v$ and $m$ mean the query, key, value and attention mask. The details of $\operatorname{Decoder}(q,k,v,m)$ are as follows. 

First, we implement the self-attention operation for query $q$ as follows:
\begin{equation}
f_1 = \operatorname{softmax}(\frac{Q_1 \cdot K_1^T}{\sqrt{C_1}})\cdot V_1
\label{eq:query_f1}
\end{equation}
Where $Q_1=W^{Q_1} \cdot q$, $K_1 = W^{K_1} \cdot q$, and $V_1 = W^{V_1} \cdot q $. $W^{Q_1}, W^{K_1}, W^{V_1}$ are learnable parameters and $C_1$ represents the channel dimension of $Q_1$. 

Then,  the cross attention is further applied to achieve the feature interaction between two modalities, which can be computed as:
\begin{equation}
O = \operatorname{softmax}(\frac{Q_2 \cdot K_2^T}{\sqrt{C_2}}+m)\cdot V_2
\label{eq:query_f2}
\end{equation}
Where $Q_2 = W^{Q_2} \cdot f_1$, $K_2 = W^{K_2} \cdot k$ and $V_2 =  W^{V_2} \cdot v$.
$W^{Q_2}, W^{K_2}, W^{V_2}$ are learnable parameters. $C_2$ is the channel dimension of $Q_2$. Note that we can set a large negative number~(\eg, $-1e^6$) in the attention mask $m$ to ignore the unnecessary area.

Finally, we modularize the above formulas ~\ref{eq:query_f1} and ~\ref{eq:query_f2} as:
\begin{equation}
O = \operatorname{Decoder}(q,k,v,m)
\label{eq:out}
\end{equation}
where $O$ indicates the output of $\operatorname{Decoder}$.



\begin{figure*}[t!]
	\centering
	\includegraphics[width=1.0\linewidth]{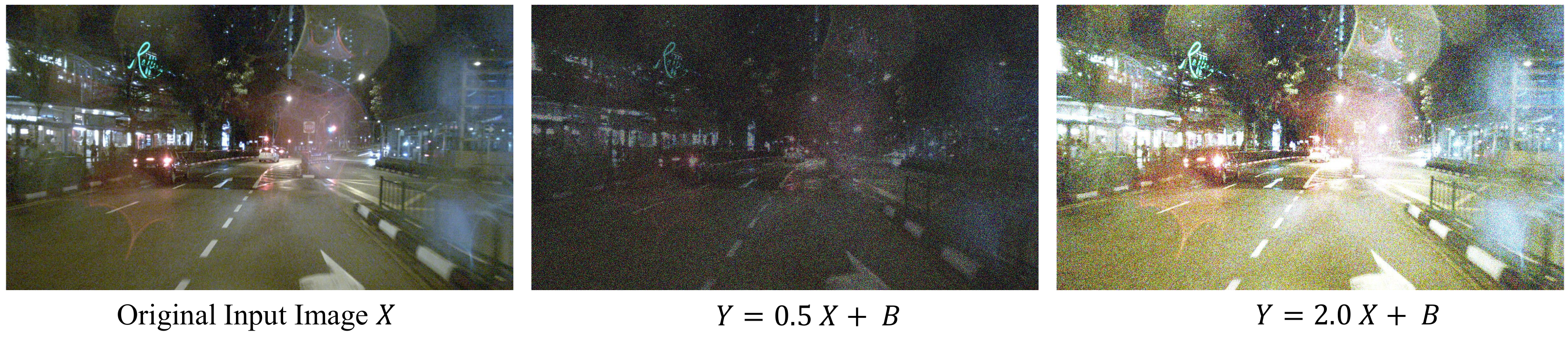}
	\caption{The visualization of introducing noisy images. The left figure is the input original image $X$, and the middle figure is processed by darkening and adding a random noisy matrix $B$ to the image $X$, which can be formulated as~$Y=0.5X+B$, and the right image is processed by lightening and adding a random noisy matrix $B$ to the image $X$, which can be computed by $Y=2.0X+B$.}\label{fig:noise_fig}
\end{figure*}

\begin{figure*}[t!]
	\centering
	\includegraphics[width=1.0\linewidth]{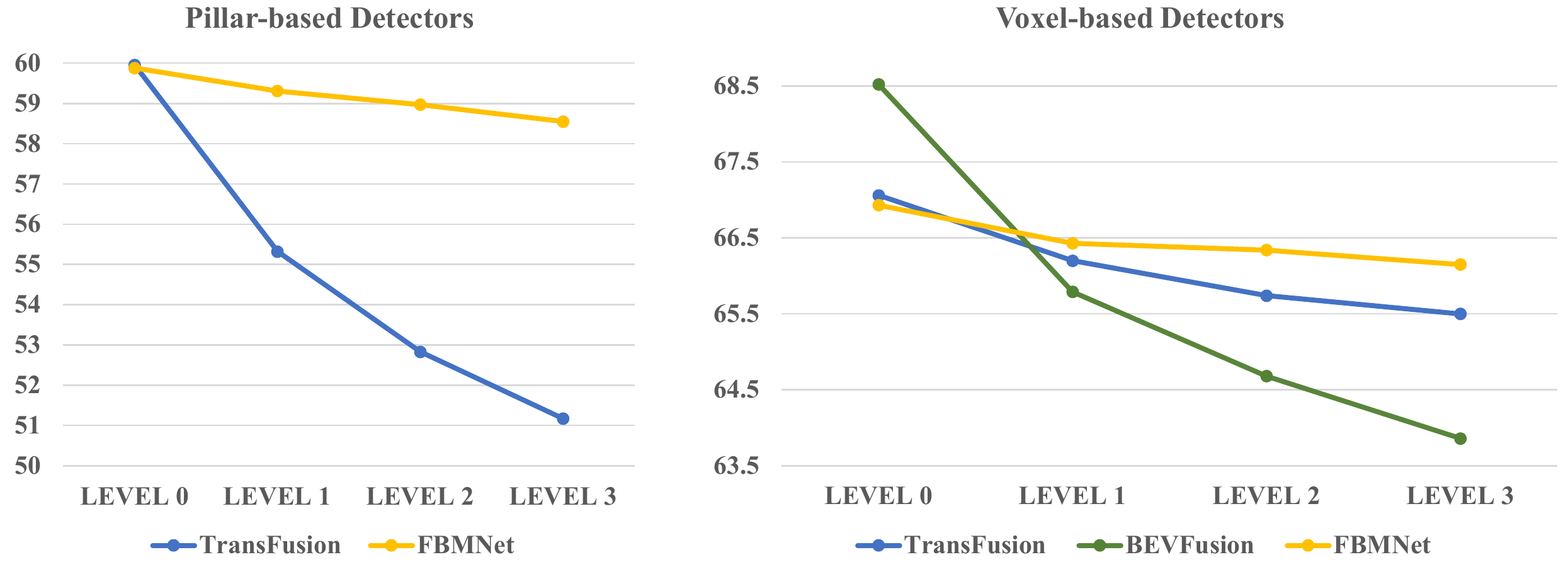}
	\caption{The results of different fusion methods under different level disturbances. LEVEL 0 means the original performance without any disturbances. We evaluate the results with mAP.}\label{fig:exp_fig}
\end{figure*}

\section{More Experiments on Robustness}
\label{sec:robust_exp}

\subsection{Inaccurate Calibration Matrix}

\begin{table}[h]
\small
\setlength{\tabcolsep}{5pt}
\resizebox{1.0\linewidth}{!}{
\centering
\begin{tabular}{l|c|c|c}
\hline
        \multirow{2}{*}{Method}  & \multicolumn{1}{c|}{\multirow{2}{*}{Backbone}}   & \multicolumn{2}{c}{Inaccurate Calibration Matrix}  \\
        \cline{3-4} 
        & \multicolumn{1}{c|}{} & \multicolumn{1}{c|}{mAP} & \multicolumn{1}{c}{NDS} \\
         \hline
        MVP~\cite{yin2021multimodal} &Pillar-based &40.61~(\robcolor{-20.11}) &51.72~(\robcolor{-14.52})   \\
        TransFusion~\cite{bai2022transfusion} &Pillar-based &42.97~(\robcolor{-16.99}) &54.63~(\robcolor{-9.71})  \\
        \mycolor{\ourMethod{}} &\mycolor{Pillar-based}  &\mycolor{\textbf{59.88}}~(\robcolor{\textbf{-0.00}}) &\mycolor{\textbf{64.93}}~(\robcolor{\textbf{-0.00}})\\
        \mycolor{Over \ourMethod{}-L} &\mycolor{Pillar-based} &\mycolor{\textbf{\redcolor{+4.84}}} &\mycolor{\textbf{\redcolor{+2.39}}} \\
        \hline
        \hline
        MVP~\cite{yin2021multimodal} &Voxel-based  &47.48~(\robcolor{-18.49}) &59.47~(\robcolor{-10.37})  \\
        TransFusion~\cite{bai2022transfusion} &Voxel-based  &64.36~(\robcolor{-2.70}) &69.30~(\robcolor{-1.46})    \\
        BEVFusion~\cite{liu2022bevfusion} &Voxel-based   &66.63~(\robcolor{-1.89}) &70.28~(\robcolor{-1.10}) ) \\
        \mycolor{\ourMethod{}} &\mycolor{Voxel-based}  &\mycolor{\textbf{66.93}}~(\robcolor{\textbf{-0.00}}) &\mycolor{\textbf{70.90}}~(\robcolor{\textbf{-0.00}})  \\
        \mycolor{Over \ourMethod{}-L} &\mycolor{Voxel-based} &\mycolor{\textbf{\redcolor{+1.95}}} &\mycolor{\textbf{\redcolor{+0.93}}}  \\
        \hline
	\end{tabular}
 }
	\caption{Robustness for inaccurate calibration matrix on nuScenes
validation set.  We evaluate the results with mAP. } 
 \label{tab:robust_calib}
\end{table}

To verify the robustness under the inaccurate calibrations of LiDAR-to-Camera, we simulate this challenging case by randomly adding a translation offset and a rotation angle to the projection matrix from LiDAR to Cameras. This may be seen as the setting of online calibration errors corresponding to offline-calibrated sensors~(the setting of misaligned sensor placement in the main paper). Specifically, the translation offset satisfies the uniform distribution of [-0.5, 0.5] meters and the rotation angle follows the uniform distribution of [-30, 30] degrees. 
The results are summarized in Table~\ref{tab:robust_calib}. 
As shown, when adding perturbance to calibration, 
all three projection-based methods suffer from performance degradation.
Besides, we observe that voxel-based fusion is generally much more robust than pillar-based  under the same method.
The reason may be that the weaker pillar-based approaches rely more on utilizing rich image semantic features to enhance LiDAR features for better performance, which will be more harmful when the fused image features are unreliable. 
Especially, it should be noted that the performance of TransFusion~\cite{bai2022transfusion} and MVP~\cite{yin2021multimodal} with PointPillars as backbones are much lower than their LiDAR-only counterparts (over 11\% drop in mAP), respectively.  
These indicate the criticality of keeping accurate calibrations during inference for existing projection-based methods.
However, benefiting from the proposed calibration-free matching strategy, \textbf{our \ourMethod{} is not affected by the projection matrix and thus maintains a steady performance.}

\subsection{Noisy Images}

\begin{table}[h]
\small
\setlength{\tabcolsep}{5pt}
\resizebox{0.95\linewidth}{!}{
\centering
\begin{tabular}{l|c|c|c}
\hline
        \multirow{2}{*}{Method}  & \multicolumn{1}{c|}{\multirow{2}{*}{Backbone}}   & \multicolumn{2}{c}{Noisy Images}  \\
        \cline{3-4} 
        & \multicolumn{1}{c|}{} & \multicolumn{1}{c|}{mAP} & \multicolumn{1}{c}{NDS} \\
         \hline
        MVP~\cite{yin2021multimodal} &Pillar-based &17.92~(\robcolor{-44.80}) &40.46~(\robcolor{-25.78})   \\
        TransFusion~\cite{bai2022transfusion} &Pillar-based &52.69~(\robcolor{-7.27}) &60.34~(\robcolor{-4.00})  \\
        \mycolor{\ourMethod{}} &\mycolor{Pillar-based}  &\mycolor{\textbf{55.29}}~(\robcolor{\textbf{-4.59}}) &\mycolor{\textbf{62.64}}~(\robcolor{\textbf{-2.29}})\\
        \mycolor{Over \ourMethod{}-L} &\mycolor{Pillar-based} &\mycolor{\textbf{\redcolor{+0.25}}} &\mycolor{\textbf{\redcolor{+0.10}}} \\
        \hline
        \hline
        MVP~\cite{yin2021multimodal} &Voxel-based  &26.89~(\robcolor{-39.08}) &47.57~(\robcolor{-22.27})  \\
        TransFusion~\cite{bai2022transfusion} &Voxel-based  &64.38~(\robcolor{-2.68}) &69.40~(\robcolor{-1.36})    \\
        BEVFusion~\cite{liu2022bevfusion} &Voxel-based   &62.25~(\robcolor{-6.27}) &68.06~(\robcolor{-3.32}) ) \\
        \mycolor{\ourMethod{}} &\mycolor{Voxel-based}  &\mycolor{\textbf{65.05}}~(\robcolor{\textbf{-1.88}}) &\mycolor{\textbf{69.95}}~(\robcolor{\textbf{-0.95}})  \\
        \mycolor{Over \ourMethod{}-L} &\mycolor{Voxel-based} &\mycolor{\textbf{\redcolor{+0.07}}} &\mycolor{\textbf{\redcolor{-0.02}}}  \\
        \hline
	\end{tabular}
 }
	\caption{Robustness for noisy images on nuScenes
validation set. We evaluate the results with mAP. } 
 \label{tab:cam_noise}
\end{table}

\begin{figure*}[t!]
	\centering
	\includegraphics[width=1\linewidth]{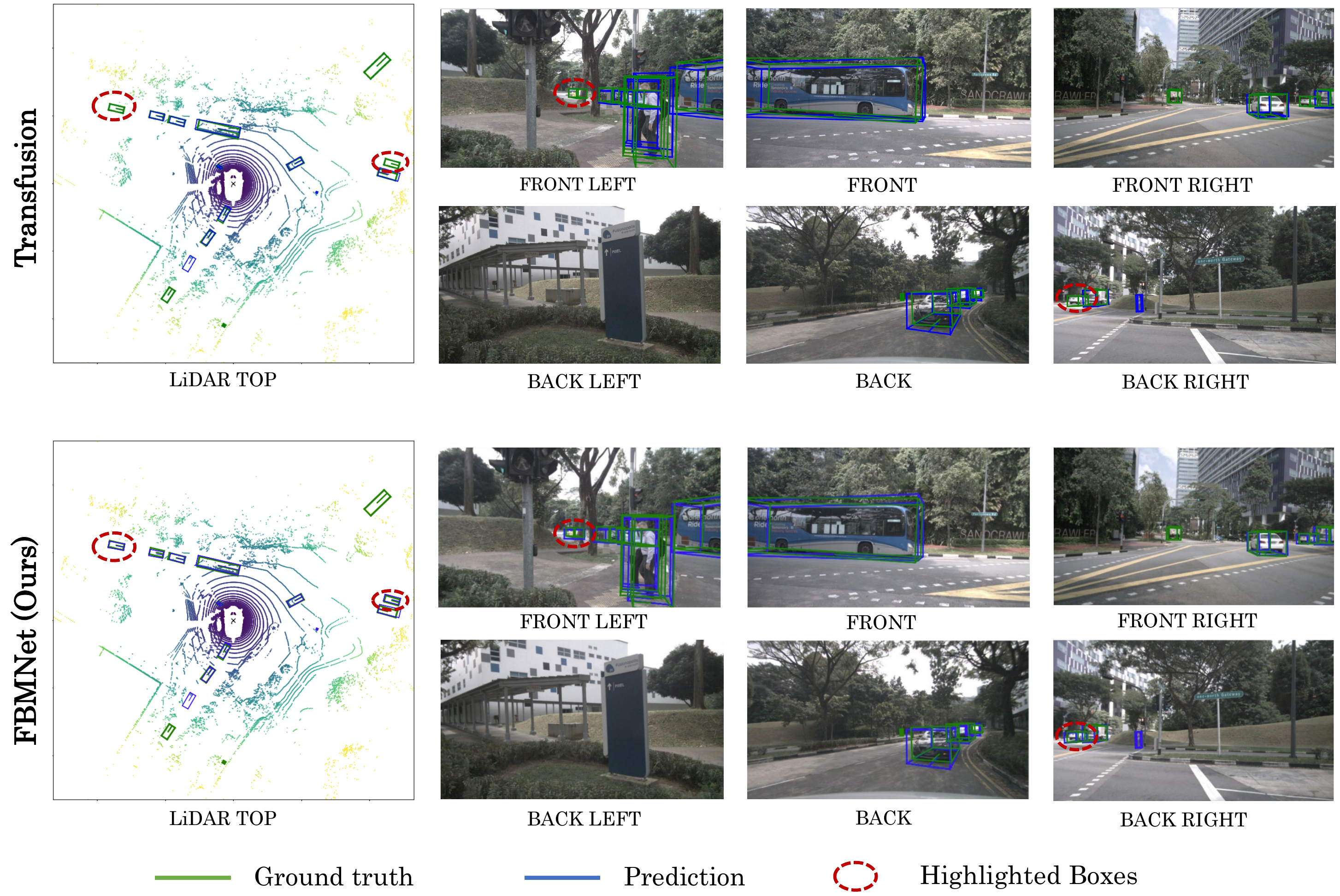}
	\caption{The visualization of our \ourMethod{} and TransFusion~\cite{bai2022transfusion} under the case of asynchronous sensors on the nuScenes validation dataset~\cite{caesar2020nuscenes}. For convenience, we also visualize the corresponding 3D boxes on 2D images.}\label{fig:vis_asyn_2s}
 \vspace{12pt}
\end{figure*}

To illustrate the robustness when the camera images are extremely disturbed by noise, we conduct experiments on the multi-modal fusion methods MVP~\cite{yin2021multimodal}, TransFusion~\cite{bai2022transfusion}, BevFusion~\cite{liu2022bevfusion}, and our \ourMethod{}. To simulate it, we process the input image by following the formula of $Y=kX+B$, where $X$ and $Y$  are the input and output RGB images, $k$ is a random constant value of 0.5 for darkening or 2.0 for lightening the input image, and $B$ is a noise matrix which satisfies the uniform distribution of (-100, 100). We present the visualization in Figure~\ref{fig:noise_fig} under the case of noisy images. Next, we provide the results of different multi-modal fusion methods under this case in Table~\ref{tab:cam_noise}. As shown, \ourMethod{} still keeps a comparable performance with the corresponding LiDAR-only detectors, which effectively proves the superiority of \ourMethod{} when images are severely disturbed. 

Besides, we also find that when three images are dropped (the setting of degenerated camera images in the main paper) or images are extremely disturbed~(the setting of noisy image), the tightly coupled fusion methods by the projection matrix even produce worse performance than the corresponding LiDAR-only detectors. Thus, to keep the system running more normally, an available strategy is to \textbf{provide an extra LiDAR-only detector for switching operations when images are extremely disturbed.}
However, our \ourMethod{} still achieves competitive results with the corresponding LiDAR-only detectors, which illustrates that our \ourMethod{} can \textbf{free up this switching operation.}


\subsection{Multiple Disturbances}

To further illustrate the superiority of \ourMethod{} for dealing with multiple disturbances, we conduct the experiments under the combined cases of asynchronous sensors and misaligned sensor placement. Considering that the sensors are usually firmly fixed~(even if there is misaligned sensor placement) after being deployed in a real scene, we mainly consider the different asynchronous times. For simplicity, we define LEVEL 1~(resp. LEVEL 2 and LEVEL 3) as the combinations of 0.08s~(resp. 0.25s and 0.50s) asynchronous times under the same  `Small' misaligned  sensor placement in our main paper. The experimental results are presented in Table~\ref{tab:robust_combined}. As shown, it can be observed that our \ourMethod{} achieves the best performance for both pillar-based and voxel-based backbones in the case of multiple disturbances, which effectively demonstrates the superiority of  \ourMethod{}. Besides, we provide a line chart as shown in Figure~\ref{fig:exp_fig} to more intuitively describe the results of different levels for fusion methods. We can observe that the performance of the advanced multi-modal detectors TransFusion~\cite{bai2022transfusion}~(pillar-based) and BEVFusion~\cite{liu2022bevfusion}~(voxel-based) has a large drop at LEVEL 1. In contrast, our FBMNet still maintains a gain of at least 1\% (voxel-based) mAP and 3\% (pillar-based) mAP over the LiDAR-only model, which illustrates \textbf{the effectiveness of fusion by box matching strategy for dealing with the above challenging cases.}

\begin{table}[t]
\small
\setlength{\tabcolsep}{3pt}
\resizebox{1.0\linewidth}{!}{
\centering
\begin{tabular}{l|c|c|c|c}
\hline
        \multirow{2}{*}{Method}  & \multicolumn{1}{c|}{\multirow{2}{*}{BK}}   & \multicolumn{3}{c}{Multiple Disturbances}  \\
        \cline{3-5} 
        & \multicolumn{1}{c|}{} & \multicolumn{1}{c|}{LEVEL 1} & \multicolumn{1}{c|}{LEVEL 2} & \multicolumn{1}{c}{LEVEL 3} \\
         \hline
        MVP~\cite{yin2021multimodal} &P &31.45~(\robcolor{{-28.08}}) &27.33~(\robcolor{{-32.20}})  &17.96~(\robcolor{{-41.57}}) \\
        TransFusion~\cite{bai2022transfusion} &P &55.32~(\robcolor{{-4.64}}) &52.83~(\robcolor{{-7.13}}) &51.17~(\robcolor{{-8.79}}) \\
        \mycolor{\ourMethod{}} &\mycolor{P}  &\mycolor{\textbf{{59.31}}}
~(\robcolor{\textbf{-0.57}}) &\mycolor{\textbf{58.97}}~(\robcolor{\textbf{-0.91}}) &\mycolor{\textbf{58.55}}~(\robcolor{\textbf{-1.33}}) \\
         \mycolor{Over \ourMethod{}-L} &\mycolor{P} &\mycolor{\textbf{\redcolor{+4.27}}} &\mycolor{\textbf{\redcolor{+3.93}}} &\mycolor{\textbf{\redcolor{+3.51}}} \\ 
        \hline
        MVP~\cite{yin2021multimodal} &V  &36.17~(\robcolor{{-29.80}}) &33.84~(\robcolor{{-32.13}})  &21.61~(\robcolor{{-44.36}}) \\
       TransFusion~\cite{bai2022transfusion} &V  &66.20~(\robcolor{{-0.86}}) & 65.74~(\robcolor{{-1.32}}) &65.50~(\robcolor{{-1.56}})  \\
        BEVFusion~\cite{liu2022bevfusion} &V   &65.79~(\robcolor{{-2.73}}) &64.68~(\robcolor{{-3.84}}) &63.86~(\robcolor{{-4.66}}) \\
        \mycolor{\ourMethod{}} &\mycolor{V}  &\mycolor{\textbf{66.43}}~(\robcolor{\textbf{-0.50}}) &\mycolor{\textbf{66.34}}(\robcolor{\textbf{-0.59}})  &\mycolor\textbf{{66.15}}(\robcolor{\textbf{-0.78}}) \\
        \mycolor{Over \ourMethod{}-L} &\mycolor{V} &\mycolor{\textbf{\redcolor{+1.45}}} &\mycolor{\textbf{\redcolor{+1.36}}} &\mycolor{\textbf{\redcolor{+1.17}}} \\
        \hline
	\end{tabular}
 }
	\caption{Robustness for the combination of inaccurate calibration, asynchronous sensors and dropped images. `P' and `V' are short for PointPillars~\cite{lang2019pointpillars} (Pillar-based) and VoxelNet~\cite{zhou2018voxelnet} (Voxel-based) as backbones. We evaluate the results with mAP.}
	\label{tab:robust_combined}
 \vspace{-8pt}
\end{table}

\begin{figure*}[t!]
	\centering
	\includegraphics[width=1\linewidth]{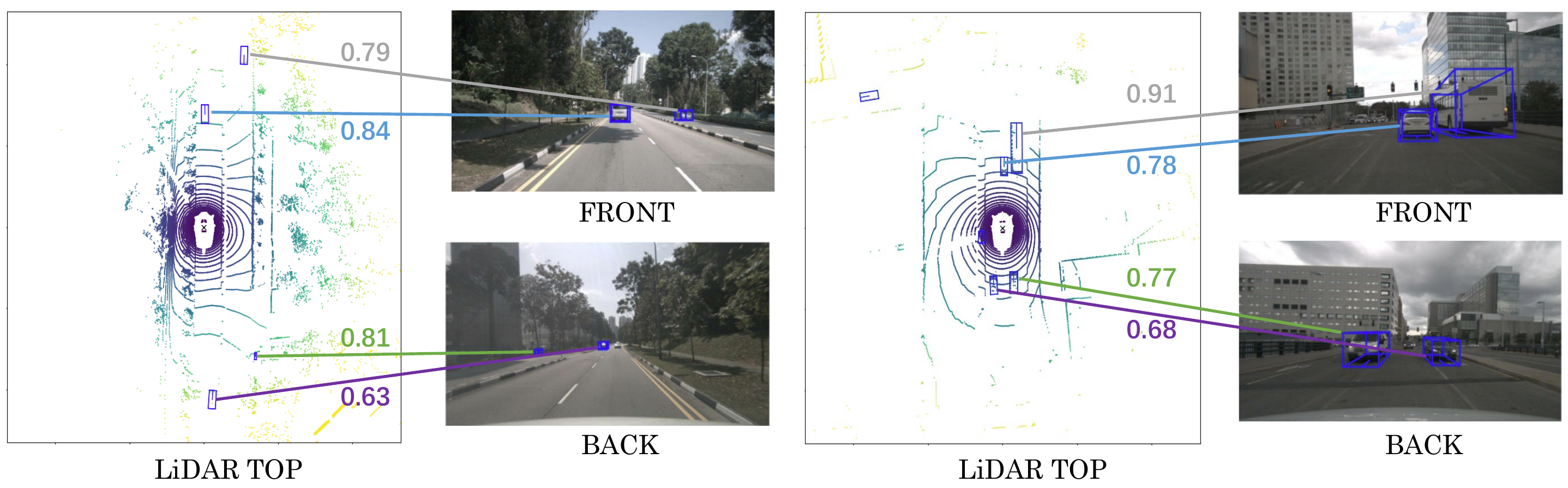}
	\caption{The visualization of the learning matching scores in the proposal-level matching module of our \ourMethod{} .}\label{fig:vis_matching_score}
\end{figure*}

\section{More Ablation Studies}
\label{sec:more_abl}

\subsection{Matching-based Fusion vs. Global Cross Attention Fusion}
\begin{table}[h!]
\small
\centering
\setlength{\tabcolsep}{12pt}
\resizebox{1.0\linewidth}{!}{
\begin{tabular}{l|cc}
\hline
         Matching  &mAP & NDS \\
         \hline
         Baseline~(TransFusion-L)~\cite{bai2022transfusion} & 64.98 &69.97 \\
         Global Cross-attention Fusion & 65.42 & 69.88  \\
         \mycolor{Matching-based Fusion~(Ours)}  & \mycolor{\textbf{66.93}} & \mycolor{\textbf{70.90}}  \\
        \hline
	\end{tabular}
 }
	\caption{Ablation study for calibration-free fusion on nuScenes \textit{validation} set. `Global Cross-attention Fusion' indicates the manner of removing the projection matrix from the original TransFusion through implementing the global cross-attention fusion inspired by ~\cite{prakash2021multi}. We evaluate the results with mAP and NDS.}
	\label{tab:abl_fusion}
\end{table}

To further illustrate the superiority of the calibration-free fusion by box matching, we remove the projection matrix from the original TransFusion~\cite{bai2022transfusion} by implementing the global cross-attention fusion inspired by ~\cite{prakash2021multi}. In detail, we set all LiDAR queries ${F}_{3d}$ as query and 2D pixel-wise image feature ${F}_{im}$ as key and value (defined in main paper). Then we encode the 2D position and 3D position information by an MLP, respectively. Next, we feed them into a DETR-like decoder~\cite{carion2020end} to achieve the calibration-free fusion. The results are summarized in Table~\ref{tab:abl_fusion}~(Line 2). As shown, the manner of global cross-attention fusion only brings an incremental improvement with mAP of 0.44\%~(65.42\% vs 64.98\%) over the LiDAR-only baseline model TransFusion-L and even produces a slight drop in NDS~(69.88\% vs 69.97\%). On the contrary, our matching-based fusion outperforms the baseline model by 1.95\% mAP and  by 0.93\% NDS, which \textbf{indicates the effectiveness of our matching-based fusion.}

\subsection{Effects of different features in the Fusion Modules}

\begin{table}[h!]
\small
\centering
\setlength{\tabcolsep}{16pt}
\resizebox{1.0\linewidth}{!}{
\begin{tabular}{l|c|c|cc}
\hline
         $O_1$ & $O_2$  & $O_3$ &mAP & NDS \\
         \hline
        --  & -- & -- & 64.98 & 69.97  \\
        $\checkmark$  & -- & -- & 66.71 & 70.71  \\
        --  & $\checkmark$ & -- & 66.53 & 70.73  \\
        --  & -- & $\checkmark$ & 66.67 & 70.82  \\
        $\checkmark$  & $\checkmark$ & -- & 66.83 & 70.79  \\
        \mycolor{$\checkmark$}  & \mycolor{$\checkmark$} & \mycolor{$\checkmark$} & \mycolor{\textbf{66.93}} & \mycolor{\textbf{70.90}}  \\
        \hline
	\end{tabular}
 }
	\caption{Ablation study for different fusion mechanisms and their combinations on nuScenes \textit{validation} set. $O_1$, $O_2$ and $O_3$ are the fused features from the formula~(7), ~(8), and ~(9), respectively. We evaluate the results with mAP and NDS.}
	\label{tab:ab_fusion}
\end{table}

Table~\ref{tab:ab_fusion} summarizes the effects of each component in the formula~(10) in the main paper and their combinations. As shown, without this module ($1^{\textit{st}}$ row), which is our baseline~(TransFusion-L~\cite{bai2022transfusion}), the performance reaches $64.98\%$ in mAP.
Integrating any of these fusion modules $O_1$, $O_2$ and $O_3$ into our baseline brings an improvement of more than 1.5\%, which indicates the image features associated through box matching are beneficial to enhance the semantic information of LiDAR features.
Combined with all components, our method achieves the mAP of 66.93\%, outperforming the baseline model of 1.95\%, which demonstrates the effectiveness of our fusion modules based on the proposed matching mechanism.

\section{Visualizations}
\label{sec:vis}

\subsection{Qualitative Comparison}

Figure~\ref{fig:vis_asyn_2s} presents a qualitative comparison of our \ourMethod{} and the popular 3D detector TransFusion~\cite{bai2022transfusion} under the case of asynchronous sensors on the nuScenes validation dataset~\cite{caesar2020nuscenes}. Specifically, we take the LiDAR's moment as the dominant and then pick the image frame advancing it by 2 seconds. As shown in Figure~\ref{fig:vis_asyn_2s}, we observe that TransFusion~\cite{bai2022transfusion} fails to detect two distant objects. The reason is that TransFusion suffers from misalignment between two modalities under the case of asynchronous sensors, which may cause misaligned image features to interfere with LiDAR features and lead to missed detections. On the contrary, \ourMethod{} can accurately identify and locate them benefiting from the proposed fusion by box matching strategy. Even in the case of asynchronous sensors,  the matching-based fusion can pick out the best-matched ROI image feature to enhance the corresponding LiDAR feature if matched ~(\eg, the matching threshold score is greater than 0.1). If not matched, FBMNet can disable the process of the LiDAR feature enhanced by the ROI image features, thus leaving the LiDAR feature undisturbed. 

\subsection{Learned Matching Results}

In Figure~\ref{fig:vis_matching_score}, we visualize the learned best matching scores (obtained by computing the maximum value of the proposal-level matching matrix $M_p$ along the row) in the proposal-level module of our \ourMethod{}. We can clearly observe that the objects of the two modalities are correctly matched even for relatively distant objects.

\end{document}